\PassOptionsToPackage{table}{xcolor}
\documentclass{article}

\usepackage{microtype}
\usepackage{graphicx}
\usepackage{subcaption}
\usepackage{lipsum}
\usepackage{graphicx}
\usepackage{amsmath}
\usepackage{amssymb}
\usepackage{subcaption}
\usepackage{booktabs}
\usepackage{algorithm}
\usepackage{algpseudocode}
\usepackage{multirow}
\usepackage{pifont} 
\usepackage{array} 
\usepackage{dsfont}
\usepackage{orcidlink}
\usepackage{enumitem}
\setlist[enumerate]{leftmargin=0mm, label=\alph*)}
\setlist[itemize]{leftmargin=3mm}
\usepackage{pifont}

\usepackage{arydshln}

\usepackage[capitalize]{cleveref}
\crefname{section}{Sec.}{Secs.}
\Crefname{section}{Section}{Sections}
\Crefname{table}{Table}{Tables}
\crefname{table}{Tab.}{Tabs.}
\usepackage[table]{xcolor}
\usepackage{colortbl}

\definecolor{KappaStarBg}{gray}{0.97}
\definecolor{KappaHalfBg}{gray}{1.0}
\definecolor{KappaStarker}{gray}{0.95}

\usepackage{tikz}

\usepackage{listings}
\usepackage{xcolor}

\lstdefinestyle{py}{
  language=Python,
  basicstyle=\ttfamily\small,
  keywordstyle=\bfseries,
  commentstyle=\itshape\color{gray},
  stringstyle=\color{black},
  showstringspaces=false,
  breaklines=true,
  columns=fullflexible,
  frame=single,
  rulecolor=\color{black},
}

\usepackage{booktabs} 

\usepackage{hyperref}

\usepackage[preprint]{icml2026}

\usepackage{amsmath}
\usepackage{amssymb}
\usepackage{mathtools}
\usepackage{amsthm}

\theoremstyle{plain}

\theoremstyle{definition}

\theoremstyle{remark}

\usepackage[textsize=tiny]{todonotes}

\icmltitlerunning{Linking Adversarial and Perturbation Robustness}

\begin{document}

\twocolumn[
\icmltitle{How Worst-Case Are Adversarial Attacks?  \\ Linking Adversarial and Perturbation Robustness}

  \icmlsetsymbol{equal}{*}
  \begin{icmlauthorlist}
    \icmlauthor{Giulio Rossolini}{yyy}
  \end{icmlauthorlist}
  \icmlaffiliation{yyy}{Department of Excellence in Robotics \& AI, Scuola Superiore Sant'Anna, Pisa, Italy}
  \icmlcorrespondingauthor{Giulio Rossolini}{giulio.rossolini@santannapisa.it}
  \icmlkeywords{Machine Learning, ICML}

  \vskip 0.3in
]

\printAffiliationsAndNotice{}  

\begin{abstract}
Adversarial attacks are widely used to identify model vulnerabilities; however, their validity as proxies for robustness to random perturbations remains debated. We ask whether an adversarial example provides a representative estimate of misprediction risk under stochastic perturbations of the same magnitude, or instead reflects an atypical worst-case event.
To address this question, we introduce a probabilistic analysis that quantifies this risk with respect to directionally biased perturbation distributions, parameterized by a concentration factor $\kappa$ that interpolates between isotropic noise and adversarial directions. Building on this, we study the limits of this connection by proposing an attack strategy designed to probe vulnerabilities in regimes that are statistically closer to uniform noise. Experiments on ImageNet and CIFAR-10 systematically benchmark multiple attacks, revealing when adversarial success meaningfully reflects robustness to perturbations and when it does not, thereby informing their use in safety-oriented robustness evaluation.
\end{abstract}

\section{Introduction}
\begin{figure}[t]
    \centering
    \begin{subfigure}{\columnwidth}
        \centering
        \includegraphics[width=\textwidth]{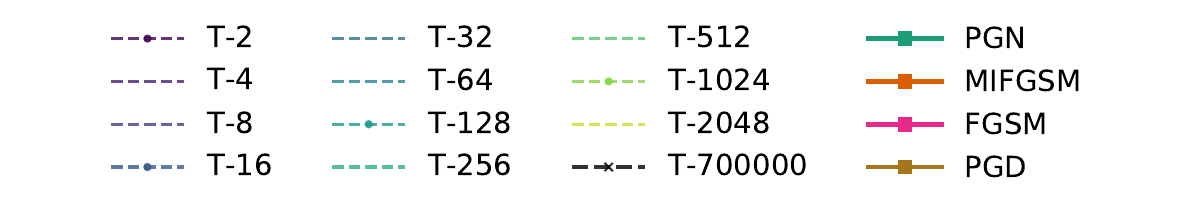}
    \end{subfigure}
    \begin{subfigure}{\columnwidth}
        \centering
        \includegraphics[width=\textwidth]{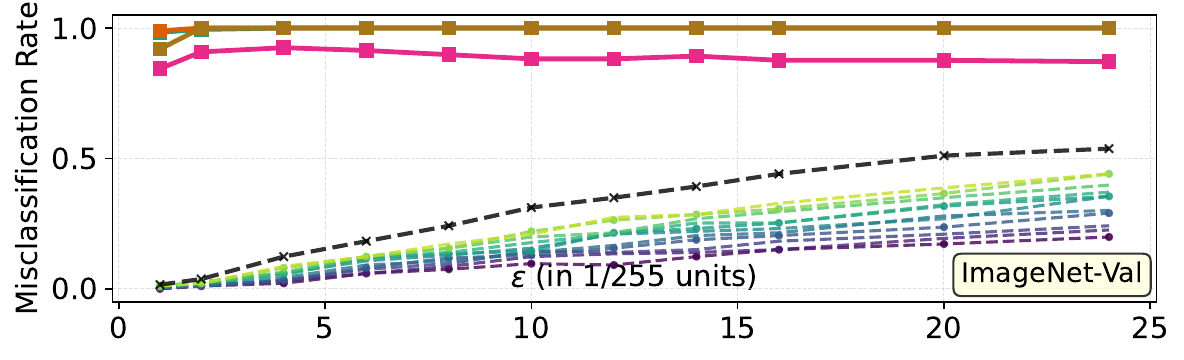}
    \end{subfigure}
    \begin{subfigure}{\columnwidth}
        \centering
        \includegraphics[width=\textwidth]{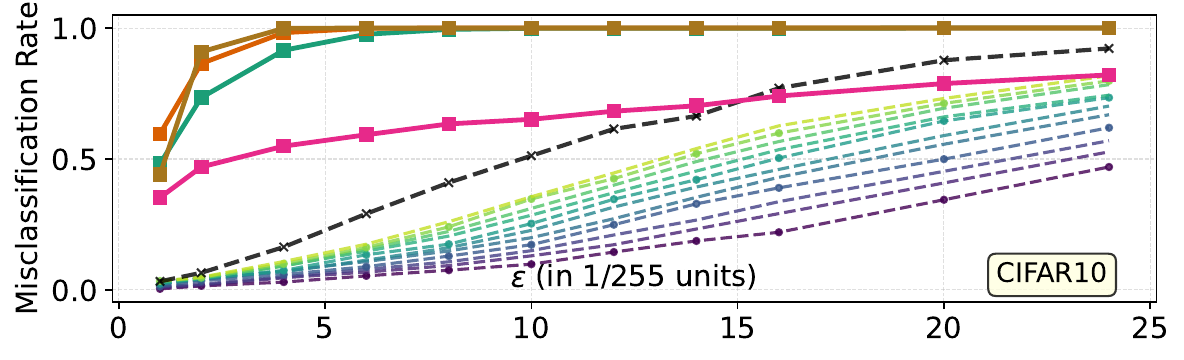}
    \end{subfigure}
    \caption{\small{Misclassification rates on ImageNet-Val and CIFAR-10 for ResNet-50 across multiple $\varepsilon$ on $\ell_\infty$. Solid lines denote adversarial attacks; dotted lines denote MC error search analysis.
 }}
    \label{f:motivation_attacks}
\end{figure}

Adversarial examples, defined as bounded perturbations intentionally crafted to alter a model prediction, have been extensively studied in the security literature~\cite{biggio2018wild}, where it is typically assumed that an attacker can deliberately manipulate digital inputs.
Beyond security, adversarial examples have also been widely adopted by the safety community as a tool to evaluate and characterize worst-case robustness to perturbations~\cite{survey_trust}.
From this perspective, the success of an adversarial attack is often treated as a conservative indicator of safety, as a sample is deemed non-robust whenever an attack succeeds.

Despite their widespread adoption, current models often exhibit strong robustness to random noise, while remaining highly vulnerable to adversarial attacks~\cite{carlini2019evaluating}. This discrepancy raises questions about whether adversarial attacks reflect robustness to statistically likely perturbations.

\noindent\textbf{Motivating the gap.}
Figure~\ref{f:motivation_attacks} illustrates this discrepancy: for clean samples that are correctly classified, we report misclassification rates obtained by applying adversarial attacks and a Monte Carlo (MC) error search based on $T$ uniformly sampled perturbations of a fixed $\varepsilon$ magnitude, where a sample is deemed misclassified if at least one of the $T$ perturbations induces an incorrect prediction.
Even for very large $T$, misclassification rate under $\varepsilon$-random perturbations remains substantially lower than that induced by adversarial attacks, becoming comparable only at large $\varepsilon$.
\footnote{$T = 700{,}000$ corresponds a probability on the order of $10^{-6}$, comparable to extremely rare catastrophic events~\cite{plait2008death}.}

In the literature, several analyses have questioned the practical relevance of adversarial attacks for safety evaluation~\cite{carlini2019evaluating, fawzi2016robustness}. While insightful, these works may discourage the use of attacks altogether, rather than clarifying when they remain informative. This motivates tools that explicitly relate adversarial success to \textit{perturbation risk}, defined as model vulnerability to stochastic perturbations of comparable magnitude.

This work introduces a probabilistic analysis of perturbation risk associated with an adversarial direction $v$ produced by an attack. We evaluate it by assessing robustness over a neighborhood of directions with fixed magnitude around $v$, modeled by a Gaussian distribution whose mean is aligned with $v$ and whose concentration is controlled by a parameter $\kappa$. Low values of $\kappa$ yield distributions close to uniform noise on the $\ell_p$-sphere, whereas larger values increasingly concentrate mass around $v$. Consequently, high attack success at low $\kappa$ indicates statistically plausible vulnerabilities, while success only at high $\kappa$ reflects localized corner cases.

Building on this analysis, we then introduce a \emph{directional noisy attack} that explicitly seeks adversarial perturbations maximizing attack success in low-$\kappa$ regimes, thereby better capturing stochastic failure regions. The attack injects controlled noise during optimization, sampled from a distribution aligned with the current adversarial direction and parameterized by a user-defined concentration.

A set of experiments illustrates the benefits of the proposed strategy, while also showing that many existing attacks achieve high success only at large $\kappa$ values, where perturbations become increasingly unrepresentative of stochastic noise of the same magnitude. These insights can help practitioners better select and complement attack evaluations for safety-oriented robustness assessment.
In summary, the main contributions of this work are:
\begin{itemize}
\item a definition of \emph{directional perturbation risk}, providing a probabilistic understanding for relating adversarial examples to stochastic perturbations of the same magnitude;
\item a \textit{directional noisy attack} optimizing adversarial success in low-$\kappa$ regimes, probing the limits of adversarial attacks as proxies for stochastic robustness;
\item a systematic experimental evaluation across multiple settings, attacks, and models, providing practical guidance for safety-oriented robustness assessment.
\end{itemize}

\section{Related work}
Adversarial robustness research has traditionally focused on worst-case perturbations to identify model vulnerabilities under security-driven threat models that explicitly assume the presence of an adversary~\cite{Szegedy14,biggio2018wild,croce2020reliable}. Beyond this scope, white-box adversarial attacks are also widely employed as safety-oriented stress tests to probe model behavior under extreme perturbations~\cite{survey_trust,ibrahum2024deep}. However, the validity of such evaluations as proxies for safety or reliability in non-adversarial settings has been repeatedly questioned, due to their limited realism and weak connection to naturally occurring perturbations~\cite{gilmer2018motivating,scher2023testingrobustnesspredictionstrained,MARCHIORIPIETROSANTI2026112412,Push_et_al}, as well as the cost and scalability challenges associated with obtaining adversarially robust models~\cite{prach20241,fawzi2018analysis}.

From a theoretical perspective, several works have questioned the interpretation and plausibility of adversarial perturbations. Geometric analyses~\cite{fawzi2016robustness} have shown that adversarial directions often exploit high-curvature regions of the decision boundary, while ~\cite{engstrom2019exploring, carlini2019evaluating} studied that model sensitivity varies substantially across perturbation directions, challenging the representativeness of worst-case attacks, as low-probability phenomena in high-dimensional input spaces.
These concerns have motivated alternative robustness evaluations based on stochastic perturbations. Randomized smoothing and related approaches assess model stability under isotropic noise and provide probabilistic robustness guarantees~\cite{cohen2019certified}. 
While effective at capturing average-case robustness, such methods rely mainly on isotropic noise, without characterizing how robustness varies across meaningful directions spaces~\cite{blum2020random,salman2019provably,tramer2020adaptive}. 

Overall, there remains substantial ambiguity regarding how adversarial attacks should be treated for safety assessment.
\cite{olivier2023many} studied the geometric abundance of adversarial perturbations via constrained directional searches, while \cite{Guo_adversarial_frontier} analyzed how adversarial behavior evolves as the perturbation magnitude increases. \cite{rice_average_case} proposed a continuum of robustness notions through functional loss norms. In contrast, this work provides a probabilistic characterization of a given adversarial direction, offering a tool for relating adversarial behavior to stochastic perturbations of the same magnitude.

\section{Background and Problem Formulation}
\noindent
Let \( f : \mathcal{X} \to \mathbb{R}^{|\mathcal{Y}|} \) be a neural network used for classification, where \( \mathcal{Y} \) denotes the set of class labels. We define
\(
\hat{y}(x) = \arg\max_{c \in \mathcal{Y}} f_c(x)
\)
as the predicted label. Let \( x \in \mathcal{X} \) be a clean input with ground-truth label \( y \in \mathcal{Y} \).
We introduce two notions of robustness and examine the challenge of bridging them.

\noindent \textit{Adversarial robustness.}
Given an adversarial attack $\mathcal{A}$, an input $x$ is
\textit{not adversarially $\varepsilon$-robust} 
if 
\begin{equation}
\exists\delta = \mathcal{A}(x, y; \varepsilon), \|\delta\|_p \le \varepsilon, \quad \textit{s.t.}  \quad \hat{y}(x+\delta) \neq y .
\label{eq:advers}
\end{equation}
This formulation relies on the existence of an adversarial perturbation $\delta$, given by an attack $\mathcal{A}$, to deem a point non-robust, without modeling the probability of such a perturbation under a statistical noise perspective.

\noindent \textit{Robustness to fixed-magnitude perturbations.}
From a general perspective, robustness to noise is a broad concept that can be formalized with respect to selected corruptions and transformations~\cite{hendrycks2019benchmarking, carvalho2025rethinking}.
To establish a direct correspondence with adversarial examples, we focus on stochastic perturbations of fixed magnitude.
We consider perturbations sampled uniformly from the $\ell_p$-sphere of radius $r$ centered at $x$.
Such a distribution can be approximated\footnote{Exact for $\ell_2$; approximate for $\ell_p$, $p \neq 2$ (see Appendix~\ref{app:lp_sampling}).} via a Gaussian
by
\(
\eta = \Pi_{\|\cdot\|_p = r}(\xi),
\)
where
\(
\xi \sim \mathcal{N}(0, I),
\)
and $\Pi_{\|\cdot\|_p = r}$ denotes projection onto the $\ell_p$-sphere of radius $r$.
Under this construction, the \textit{perturbation risk} of a correctly classified sample $x$
(i.e., $\hat{y}(x) = y$) is defined as
\begin{equation}
\Pr_{\xi \sim \mathcal{N}(0, I)}
\big[
\hat{y}(x + \Pi_{\|\cdot\|_p = r}(\xi)) \neq y
\big].
\label{eq:noisy_rob}
\end{equation}
While perturbation risk provides a statistically meaningful notion for safety assessment, exhaustively estimating it in practice via Monte Carlo sampling may require prohibitive evaluation time~\cite{cohen2019certified}.
Consequently, adversarial attacks are treated as pointwise robustness surrogates, even though they capture fundamentally different failure notions: the existence of an adversarial perturbation $\delta$ only certifies the presence of a specific error direction and does not imply that such a direction is representative under stochastic perturbations of the same magnitude (Eq.~\ref{eq:noisy_rob} for $r = \|\delta\|_p$). As a result, the common empirical practice of inferring perturbation risk behavior from adversarial success (Eq.~\ref{eq:advers}) is generally unjustified, as any potential relationship critically depends on the properties of the attack $\mathcal{A}$ and on the geometry of the model loss landscape~\cite{li2018visualizing}.
This motivates the need for tools that quantify how adversarial directions relate to statistically plausible perturbations.\footnote{We use the same $p$-norm for both adversarial attacks and perturbation risk; this choice is not restrictive (see Appendix~\ref{app:lp_sampling}).}

\section{Directional Perturbation Risk Metric}
\label{sec:methodology}
We study whether the unit adversarial direction
\(
v \;=\; \frac{\delta}{\|\delta\|_2} 
\)
points toward a representative noisy failure region on a $\ell_p$-sphere of a radius $r$, thereby linking adversarial and perturbation robustness.
To explore this, rather than focusing on isolated perturbations or uniform ones, we aim to characterize whether misclassification persists under stochastic perturbations that are biased toward $v$.
Specifically, we generate perturbations by sampling from a Gaussian
distribution whose mean is shifted along the direction $v$ and then projecting onto the $r$-sphere:
\begin{equation}
\eta = \Pi_{\|\cdot\|_p = r}(\xi),
\qquad
\xi \sim \mathcal{N}(\kappa v, I),
\label{eq:dir_gauss}
\end{equation}
where $\kappa \ge 0$ controls the strength of the directional bias: $\kappa = 0$, this reduces to uniform noise on the sphere (as in Eq.\ref{eq:noisy_rob}), while larger values of $\kappa$ progressively concentrate probability mass around the direction $v$.
We thus define the \emph{directional perturbation risk} of a direction $v$ as
\begin{equation}
\mathcal{R}(x; v, \kappa, r)
\;=\;
\Pr_{\xi \sim \mathcal{N}(\kappa v, I)}
\big[
\hat{y}(x + \Pi_{\|\cdot\|_p = r}(\xi)) \neq y
\big].
\label{eq:dir_noisy_risk}
\end{equation}
In practice, $\mathcal{R}$ can be estimated numerically as the misclassification rate over $n$ samples.
For direct comparison with an adversarial perturbation \(\delta\), we consider \(r = \|\delta\|_p\).
By varying \(\kappa\) as an evaluation parameter, we can analyze different levels of perturbation risk, ranging from uniform noise on the $\ell_p$-sphere to perturbations concentrated around the adversarial direction \(v\) as \(\kappa \to \infty\), which progressively resemble the original attack success rate.

\begin{figure}[t]
    \centering
    \begin{subfigure}{0.32\columnwidth}
        \centering
        \includegraphics[width=\textwidth]{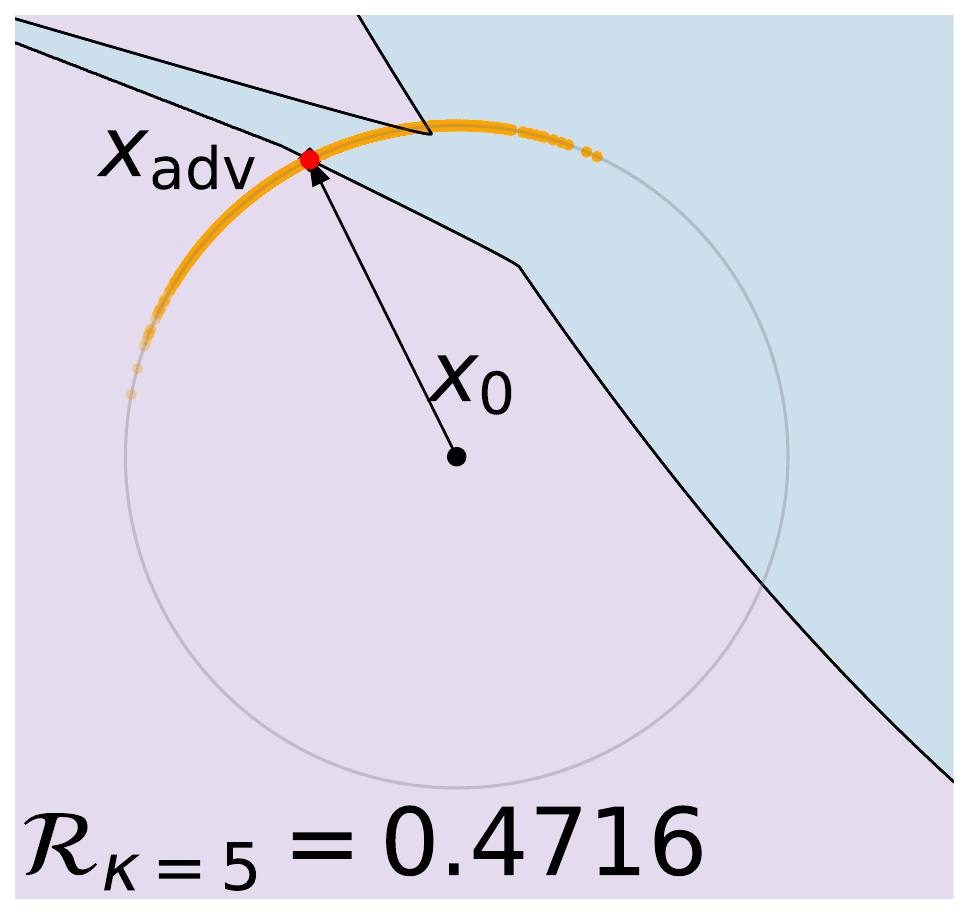}
        \caption{}
        \label{demo_a}
    \end{subfigure}
    \begin{subfigure}{0.32\columnwidth}
        \centering
        \includegraphics[width=\textwidth]
        {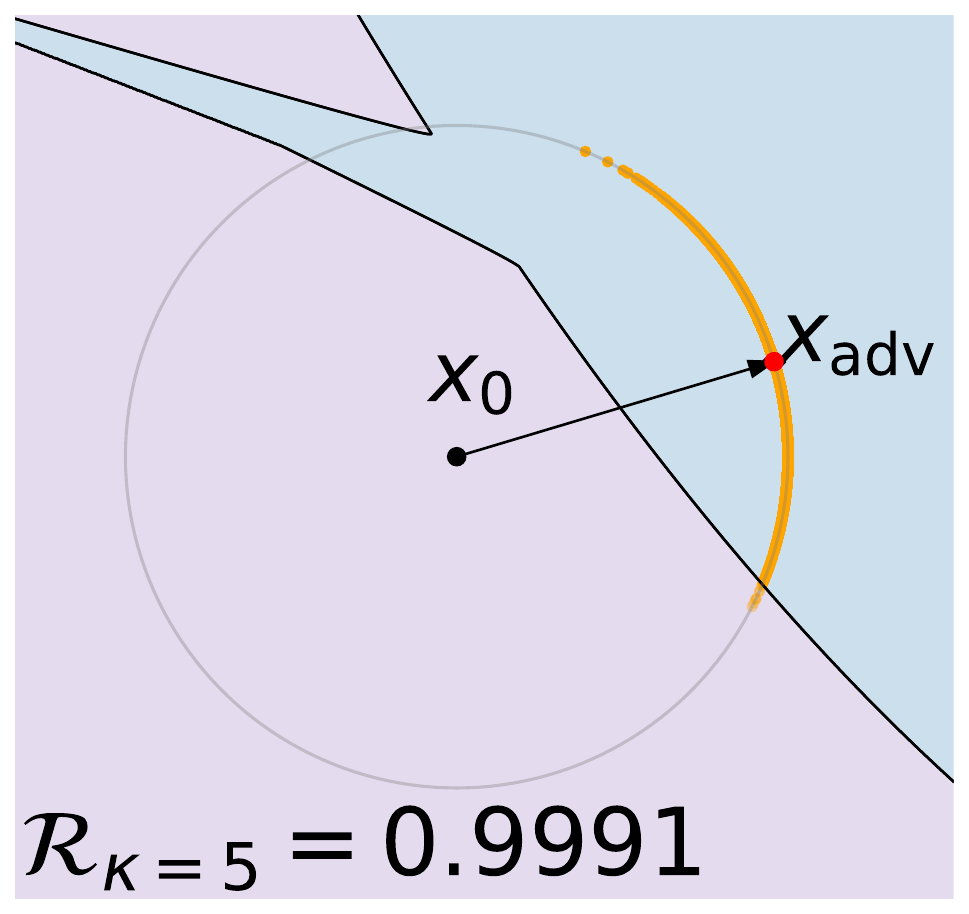}
        \caption{}
        \label{demo_b}
    \end{subfigure}
    \begin{subfigure}{0.32\columnwidth}
        \centering
        \includegraphics[width=\textwidth]{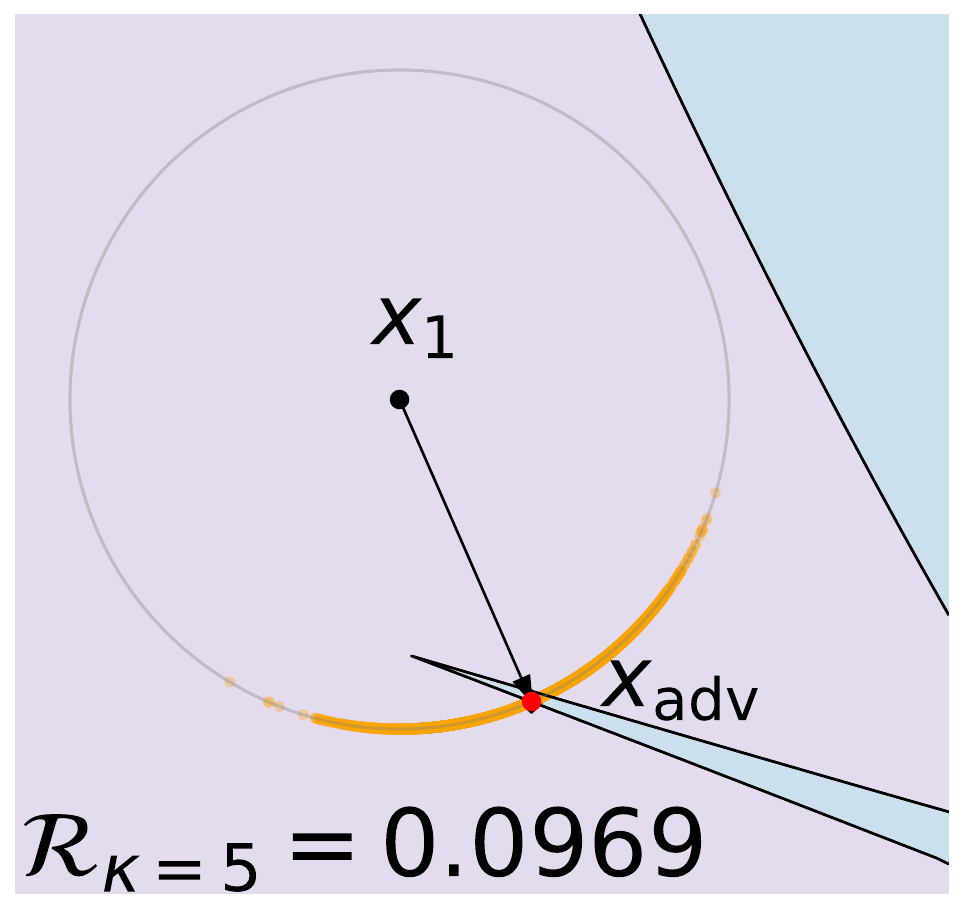}
        \caption{}
        \label{demo_c}
    \end{subfigure}
    \caption{\small{Illustrative examples. (a) and (b) compare two adversarial directions for the same input $x_0$ that differ in their directional perturbation risk, computed with $\kappa = 5$. (c) depicts a corner-case adversarial direction with low perturbation risk.}}
    \label{f:demo_attack}
\end{figure}

\paragraph{Geometric Intuition.}
Figure~\ref{f:demo_attack} illustrates the proposed metric in a simplified two-dimensional setting.
Fig.~\ref{demo_a} and~ Fig.\ref{demo_b} show two adversarial examples on the same $\ell_2$-sphere centered at $x_0$ that induce misclassification but differ in their directional perturbation risk $\mathcal{R}$.
Although both attacks are successful, the example in Fig.~\ref{demo_b} yields a much higher $\mathcal{R}$ (here with $\kappa=5$), indicating that a larger neighborhood of nearby perturbations remains misclassified.
In contrast, the adversarial direction in Fig.~\ref{demo_a} corresponds to a more localized failure region, representing a lower perturbation risk.
Figure~\ref{demo_c} considers a different input point $x_1$, where misclassification occurs only along a narrow, corner region of the decision boundary.
Despite being adversarial, this example yields a very low perturbation risk for $\kappa=5$, revealing its highly localized and non-representative nature.

\paragraph{On the role of $\kappa$.}
From a practical perspective, to match a stochastic scenario, $\kappa$ should be chosen such that the resulting perturbations remain statistically close
to uniform noise on the $\ell_p$-sphere.
For $p=2$, consider $D$ the input dimensions and write $\xi = \kappa v + z$, with $z \sim \mathcal{N}(0,I)$, by concentration of measure, the norm of the random component satisfies $\|z\|_2 \approx \sqrt{D}$ with high probability. As a consequence, the directional shift $\kappa v$ remains negligible compared to the
$z$ as long as $\kappa \ll \sqrt{D}$, while for $\kappa \gtrsim \sqrt{D}$ the perturbations increasingly concentrate around $v$.
However, although $\sqrt{D}$ emerges as a natural scale from a norm-based perspective,
in practice, values of $\kappa$ on the order of $\sqrt{D}$ already induce a strongly
directional perturbation distribution, due to the rapid concentration of probability mass
around the mean direction in high-dimensional spaces. As a result, we propose $\kappa^* = D^{1/4}$ as an empirical yet more conservative reference point. This choice lies well within a stochastic regime, while still introducing a measurable directional preference after projection. We further justify this choice experimentally in Section~\ref{sec:correlaton_kappa} for both $p=2$ and $p=\infty$ perturbations. Additional analyses are in Appendix~\ref{app:kappa_scaling}.

\section{Directional Noisy Attack}
\label{sec:directional_attack}
\begin{algorithm}[t]
\caption{DN-Attack}
\label{alg:vmf_pgd}
\begin{algorithmic}[1]
\Require
$x$ , $y$, 
$f$ , $\mathcal{L}$,
$\varepsilon$, $p$,
$T$, $N$,
$\alpha$, $\mu$,
$\kappa_\text{adv}$ 

\State $x_{\text{adv}} \gets x$, $g \gets 0$
\For{$t = 1$ to $T$}
    \State $\bar{g} \gets 0$
    \For{$i = 1$ to $N$}
        \If{$t == 1$} 
            \State \#\textit{Add random noisy in $\varepsilon$}
            \State Sample $\eta \sim \texttt{rand}$ in $\mathcal{B}_p(0,\varepsilon))$
            \State $x_i \gets \mathrm{clip}_{[0,1]}(x + \eta)$
        \Else
            \State \#\textit{Add directional noisy from $\delta$}
            \State $\delta \gets x_{\text{adv}} - x$
            \State $v \gets \delta/\|\delta\|_2$
            \State Sample $\xi \sim \mathcal{N}(\kappa_{\text{adv}} v, I)$
            \State $\eta \gets \Pi_{\|\cdot\|_p = \|\delta\|_p}(\xi)$
            \State $x_i \gets \mathrm{clip}_{[0,1]}(x + \eta)$
            
        \EndIf
        \State $\bar{g} \gets \bar{g} + \nabla_x \mathcal{L}(f(x_i), y)$
    \EndFor
    \State $g \gets \mu g + \mathrm{Normalize_p}(\bar{g})$
    \State $x_{\text{adv}} \gets \Pi_{\mathcal{B}_p(x,\varepsilon)}(x_{\text{adv}} + \alpha\, g)$
    \State $x_{\text{adv}} \gets \mathrm{clip}_{[0,1]}(x_{\text{adv}})$
\EndFor
\State \Return $x_{\text{adv}}$

\end{algorithmic}
\end{algorithm}

Well-known adversarial strategies, such as PGD~\cite{pgd_attack}, DeepFool~\cite{Moosavi16_deepfool}, and CW~\cite{Carlini017}, iteratively follow the local steepest-ascent direction of the task objective to maximize $\mathcal{L}$ in untargeted settings, i.e
\(
\max_{\lVert \delta \rVert_p \leq \varepsilon}
\mathcal{L}\!\left(f(x + \delta), y\right).
\)

While effective at identifying worst-case perturbations, these methods provide no probabilistic characterization of whether the resulting adversarial direction corresponds to a broad failure region or merely an isolated point \cite{fawzi2016robustness}.
To favor adversarial directions associated with high-probability failure regions, we follow an alternative objective that directly maximizes the proposed directional perturbation risk at a selected concentration ${\kappa}_{\text{adv}}$
\begin{equation}
\max_{\lVert \delta \rVert_p \leq \varepsilon}
\mathcal{R}\!\left(
x,\;
v,\;
\kappa_{\text{adv}},\;
r
\right),
\label{eq:optimization_problem}
\end{equation}
where $v = \delta / \lVert \delta \rVert_2$, and $ r = \lVert \delta \rVert_p$.
To implement this, we propose an iterative attack strategy (Algorithm~\ref{alg:vmf_pgd}) that explicitly optimizes over a neighborhood of directions around the current adversarial direction, thereby increasing the probability of encountering adversarial perturbations corresponding to statistically representative noisy regions.
In particular, let \(\delta^t = x^t_{\text{adv}} - x\) denote the perturbation at iteration \(t\), and define the normalized adversarial direction for $t > 1$, as
\(
    v^t = \frac{\delta^t}{\|\delta^t\|_2}.
\)
At the first iteration ($t=1$), since \(\delta^t = 0\), the attack estimates the gradient by sampling \(N\) random points within the \(\mathcal{B}_p(x,\varepsilon)\), as in~\cite{ge2023boosting}. This initialization avoids biasing the attack toward an arbitrary direction before a meaningful adversarial direction is identified.
For subsequent iterations (\(t > 1\)), gradients are computed by exploring ascent directions sampled around the current adversarial direction \(v^t\). 
Specifically, for $N$ times, we sample a random vector
\(
    \xi \sim \mathcal{N}(\kappa_{\text{adv}} v^t, I),
\)
which induces a directionally biased distribution centered at $v^t$ with concentration
$\kappa_{\text{adv}}$.
The sampled vector is then projected onto the $\ell_p$-sphere with radius
$\|\delta^t\|_p$, yielding the perturbation
\(
    \eta = \Pi_{\|\cdot\|_p = \|\delta^t\|_p}(\xi).
\)
This construction ensures that each perturbation matches the current magnitude of the
adversarial iterate while preserving a controlled directional bias.
Following this procedure, for $t > 1$ the update, with step size $\alpha$ and momentum $\mu$, is guided by the gradient
\begin{equation}
    \bar{g}^t
    \;\approx\;
    \mathbb{E}_{\xi \sim \mathcal{N}(\kappa_{\text{adv}} v^t, I)}
    \Bigl[
        \nabla_x \ell\bigl(
            f(x + \eta),
            y
        \bigr)
    \Bigr], 
\end{equation}
which captures ascent directions that are adversarially meaningful across a coherent cone of nearby directions.

Here, the parameter $\kappa_{\text{adv}}$ controls the trade-off between directional
exploitation and exploration.
Considering the limit $\kappa_{\text{adv}} \rightarrow \infty$, the sampling distribution collapses around $v^t$, and the attack behavior approaches that of standard PGD.
Conversely, small values of $\kappa_{\text{adv}}$ lead to nearly isotropic sampling, which
may reduce effectiveness in worst-case regimes.
In line with the goal of this work, we therefore focus on low values of
$\kappa_{\text{adv}}$, as the proposed DN attack is not intended to replace state-of-the-art techniques for worst-case robustness, but rather to expose statistically representative failure regions. Ablation studies in Section~\ref{sec:ablation} and Appendix \ref{app:cifar10_results} analyze this trade-off in detail.

\begin{table*}[h]
\centering
\caption{\small{Directional perturbation risk under $\ell_\infty$ attacks ($\varepsilon \in \{4/255, 8/255\}$) and $\ell_2$ attacks ($\varepsilon \in \{1.0, 2.0\}$). 
$\bar{\mathcal{R}}_{\kappa^*}$ denotes the average directional perturbation risk on the ImageNet-Val evaluated at $\kappa^* = D^{1/4}$, while $\kappa_{0.25}$ denotes the smallest concentration parameter $\kappa$ estimated for which $\bar{\mathcal{R}} \ge 0.25$. 
ASR denotes the standard attack success rate, reported for comparison.
}}
\label{table:imagenet_attacks}
\resizebox{0.95\textwidth}{!}{
\begin{tabular}{ll|c|c|c|c|c|c|c|c|c|c|c|c}

\toprule
\cmidrule(lr){3-8}\cmidrule(lr){9-14}
& & \multicolumn{3}{c|}{$\ell_\infty$,  $\varepsilon = 4/255$} & \multicolumn{3}{c|}{$\ell_\infty$,  $\varepsilon = 8/255$}
& \multicolumn{3}{c|}{$\ell_2$,  $\varepsilon = 1.0$} & \multicolumn{3}{c}{$\ell_2$, $\varepsilon = 2.0$} \\
\midrule

Model & Attack
& $k_{\bar{\mathcal{R}}={0.25}}\!\downarrow$ & $\bar{\mathcal{R}}_{\kappa^*}\!\uparrow$ & $\textit{ASR}\!\uparrow$
& $k_{\bar{\mathcal{R}}={0.25}}\!\downarrow$ & $\bar{\mathcal{R}}_{\kappa^*}\!\uparrow$  & $\textit{ASR}\!\uparrow$
& $k_{\bar{\mathcal{R}}={0.25}}\!\downarrow$ & $\bar{\mathcal{R}}_{\kappa^*}\!\uparrow$  & $\textit{ASR}\!\uparrow$
& $k_{\bar{\mathcal{R}}={0.25}}\!\downarrow$ & $\bar{\mathcal{R}}_{\kappa^*}\!\uparrow$  & $\textit{ASR}\!\uparrow$ \\
\midrule


{ResNet-50} 
& DN    
& \cellcolor{KappaStarker}\textbf{41}  & \cellcolor{KappaStarBg}\textbf{0.106} & \cellcolor{KappaHalfBg}{0.990}
& \cellcolor{KappaStarker}\textbf{26}  & \cellcolor{KappaStarBg}\textbf{0.178} & \cellcolor{KappaHalfBg}{0.995}
& \cellcolor{KappaStarker}\textbf{31}  & \cellcolor{KappaStarBg}\textbf{0.128} & \cellcolor{KappaHalfBg}{0.957}
& \cellcolor{KappaStarker}\textbf{19}  & \cellcolor{KappaStarBg}\textbf{0.266} & \cellcolor{KappaHalfBg}{0.973} \\

& FGSM    
& \cellcolor{KappaStarker}\underline{52} & \cellcolor{KappaStarBg}\underline{0.094} & \cellcolor{KappaHalfBg}0.929
& \cellcolor{KappaStarker}\underline{33} & \cellcolor{KappaStarBg}\underline{0.144} & \cellcolor{KappaHalfBg}0.909
& \cellcolor{KappaStarker}{39} & \cellcolor{KappaStarBg}\underline{0.117} & \cellcolor{KappaHalfBg}0.801
& \cellcolor{KappaStarker}{25} & \cellcolor{KappaStarBg}\underline{0.200} & \cellcolor{KappaHalfBg}0.883 \\

& MI-FGSM 
& \cellcolor{KappaStarker}54 & \cellcolor{KappaStarBg}0.079 & \cellcolor{KappaHalfBg}{0.999}
& \cellcolor{KappaStarker}37 & \cellcolor{KappaStarBg}0.118 & \cellcolor{KappaHalfBg}\textbf{1.000}
& \cellcolor{KappaStarker}\underline{34} & \cellcolor{KappaStarBg}0.113 & \cellcolor{KappaHalfBg}{0.979}
& \cellcolor{KappaStarker}\underline{24} & \cellcolor{KappaStarBg}0.197 & \cellcolor{KappaHalfBg}{0.997} \\

& PGD-5   
& \cellcolor{KappaStarker}153 & \cellcolor{KappaStarBg}0.026 & \cellcolor{KappaHalfBg}0.999
& \cellcolor{KappaStarker}135 & \cellcolor{KappaStarBg}0.036 & \cellcolor{KappaHalfBg}\textbf{1.000}
& \cellcolor{KappaStarker}69  & \cellcolor{KappaStarBg}0.056 & \cellcolor{KappaHalfBg}0.966
& \cellcolor{KappaStarker}75  & \cellcolor{KappaStarBg}0.054 & \cellcolor{KappaHalfBg}0.970 \\

& PGD-20  
& \cellcolor{KappaStarker}244 & \cellcolor{KappaStarBg}0.015 & \cellcolor{KappaHalfBg}\textbf{1.000}
& \cellcolor{KappaStarker}165 & \cellcolor{KappaStarBg}0.031 & \cellcolor{KappaHalfBg}\textbf{1.000}
& \cellcolor{KappaStarker}68  & \cellcolor{KappaStarBg}0.053 & \cellcolor{KappaHalfBg}\textbf{0.999}
& \cellcolor{KappaStarker}56  & \cellcolor{KappaStarBg}0.069 & \cellcolor{KappaHalfBg}\textbf{1.000} \\

& PGN     
&\cellcolor{KappaStarker} 67 & \cellcolor{KappaStarBg}0.061 & \cellcolor{KappaHalfBg}\textbf{1.000}
& \cellcolor{KappaStarker}52 & \cellcolor{KappaStarBg}0.080 & \cellcolor{KappaHalfBg}\textbf{1.000}
& \cellcolor{KappaStarker}50 & \cellcolor{KappaStarBg}0.071 & \cellcolor{KappaHalfBg}\underline{0.997}
& \cellcolor{KappaStarker}27 & \cellcolor{KappaStarBg}0.165 & \cellcolor{KappaHalfBg}\underline{0.999} \\

\midrule

{VGG-16} 
& DN      
& \cellcolor{KappaStarker}\textbf{31} & \cellcolor{KappaStarBg}\textbf{0.148} & \cellcolor{KappaHalfBg}0.995
& \cellcolor{KappaStarker}\textbf{21} & \cellcolor{KappaStarBg}\textbf{0.234} & \cellcolor{KappaHalfBg}0.999
& \cellcolor{KappaStarker}\textbf{24} & \cellcolor{KappaStarBg}\textbf{0.199} & \cellcolor{KappaHalfBg}0.993
& \cellcolor{KappaStarker}\textbf{14} & \cellcolor{KappaStarBg}\textbf{0.360} & \cellcolor{KappaHalfBg}0.994 \\

& FGSM    
& \cellcolor{KappaStarker}\underline{38} & \cellcolor{KappaStarBg}\underline{0.130} & \cellcolor{KappaHalfBg}0.977
& \cellcolor{KappaStarker}\underline{26} & \cellcolor{KappaStarBg}\underline{0.186} & \cellcolor{KappaHalfBg}0.962
& \cellcolor{KappaStarker}\underline{27} & \cellcolor{KappaStarBg}\underline{0.181} & \cellcolor{KappaHalfBg}0.950
& \cellcolor{KappaStarker}\underline{17} & \cellcolor{KappaStarBg}\underline{0.292} & \cellcolor{KappaHalfBg}0.973 \\

& MI-FGSM 
& \cellcolor{KappaStarker}42 & \cellcolor{KappaStarBg}0.105 & \cellcolor{KappaHalfBg}\textbf{0.999}
& \cellcolor{KappaStarker}31 & \cellcolor{KappaStarBg}0.151 & \cellcolor{KappaHalfBg}0.999
& \cellcolor{KappaStarker}\underline{27} & \cellcolor{KappaStarBg}0.171 & \cellcolor{KappaHalfBg}0.996
& \cellcolor{KappaStarker}19 & \cellcolor{KappaStarBg}0.268 & \cellcolor{KappaHalfBg}0.997 \\

& PGD-5   
& \cellcolor{KappaStarker}129 & \cellcolor{KappaStarBg}0.030 & \cellcolor{KappaHalfBg}\textbf{0.999}
& \cellcolor{KappaStarker}125 & \cellcolor{KappaStarBg}0.041 & \cellcolor{KappaHalfBg}\textbf{1.000}
& \cellcolor{KappaStarker}55  & \cellcolor{KappaStarBg}0.072 & \cellcolor{KappaHalfBg}0.995
& \cellcolor{KappaStarker}63  & \cellcolor{KappaStarBg}0.066 & \cellcolor{KappaHalfBg}0.994 \\

& PGD-20  
& \cellcolor{KappaStarker}203 & \cellcolor{KappaStarBg}0.019 & \cellcolor{KappaHalfBg}\textbf{0.999}
& \cellcolor{KappaStarker}152 & \cellcolor{KappaStarBg}0.036 & \cellcolor{KappaHalfBg}\textbf{1.000}
& \cellcolor{KappaStarker}60  & \cellcolor{KappaStarBg}0.061 & \cellcolor{KappaHalfBg}\textbf{0.999}
& \cellcolor{KappaStarker}50  & \cellcolor{KappaStarBg}0.082 & \cellcolor{KappaHalfBg}\textbf{0.999} \\

& PGN     
& \cellcolor{KappaStarker}55 & \cellcolor{KappaStarBg}0.073 & \cellcolor{KappaHalfBg}\textbf{0.999}
& \cellcolor{KappaStarker}45 & \cellcolor{KappaStarBg}0.101 & \cellcolor{KappaHalfBg}0.999
& \cellcolor{KappaStarker}42 & \cellcolor{KappaStarBg}0.097 & \cellcolor{KappaHalfBg}\textbf{0.999}
& \cellcolor{KappaStarker}22 & \cellcolor{KappaStarBg}0.223 & \cellcolor{KappaHalfBg}\underline{0.998} \\

\midrule

{ViT-B} 
& DN      
& \cellcolor{KappaStarker}\textbf{83} & \cellcolor{KappaStarBg}\textbf{0.045} & \cellcolor{KappaHalfBg}0.782
& \cellcolor{KappaStarker}\textbf{44} & \cellcolor{KappaStarBg}\textbf{0.082} & \cellcolor{KappaHalfBg}0.878
& \cellcolor{KappaStarker}\textbf{61} & \cellcolor{KappaStarBg}\textbf{0.061} & \cellcolor{KappaHalfBg}0.544
& \cellcolor{KappaStarker}\textbf{31} & \cellcolor{KappaStarBg}\textbf{0.140} & \cellcolor{KappaHalfBg}0.595 \\

& FGSM    
& \cellcolor{KappaStarker}114 & \cellcolor{KappaStarBg}\textbf{0.045} & \cellcolor{KappaHalfBg}0.645
& \cellcolor{KappaStarker}\underline{59}  & \cellcolor{KappaStarBg}\underline{0.081} & \cellcolor{KappaHalfBg}0.701
& \cellcolor{KappaStarker}110 & \cellcolor{KappaStarBg}\underline{0.062} & \cellcolor{KappaHalfBg}0.404
& \cellcolor{KappaStarker}51  & \cellcolor{KappaStarBg}\underline{0.112} & \cellcolor{KappaHalfBg}0.461 \\

& MI-FGSM 
& \cellcolor{KappaStarker}\underline{100} & \cellcolor{KappaStarBg}\underline{0.038} & \cellcolor{KappaHalfBg}0.955
& \cellcolor{KappaStarker}\underline{59}  & \cellcolor{KappaStarBg}0.065 & \cellcolor{KappaHalfBg}0.998
& \cellcolor{KappaStarker}\underline{67}  & \cellcolor{KappaStarBg}0.058 & \cellcolor{KappaHalfBg}0.621
& \cellcolor{KappaStarker}\underline{37}  & \cellcolor{KappaStarBg}0.106 & \cellcolor{KappaHalfBg}0.772 \\

& PGD-5   
& \cellcolor{KappaStarker}230 & \cellcolor{KappaStarBg}0.017 & \cellcolor{KappaHalfBg}0.962
& \cellcolor{KappaStarker}152 & \cellcolor{KappaStarBg}0.025 & \cellcolor{KappaHalfBg}0.977
& \cellcolor{KappaStarker}115 & \cellcolor{KappaStarBg}0.034 & \cellcolor{KappaHalfBg}0.550
& \cellcolor{KappaStarker}105 & \cellcolor{KappaStarBg}0.033 & \cellcolor{KappaHalfBg}0.555 \\

& PGD-20  
& \cellcolor{KappaStarker}379 & \cellcolor{KappaStarBg}0.009 & \cellcolor{KappaHalfBg}\textbf{0.997}
& \cellcolor{KappaStarker}202 & \cellcolor{KappaStarBg}0.019 & \cellcolor{KappaHalfBg}\textbf{1.000}
& \cellcolor{KappaStarker}116 & \cellcolor{KappaStarBg}0.030 & \cellcolor{KappaHalfBg}\textbf{0.866}
& \cellcolor{KappaStarker}78  & \cellcolor{KappaStarBg}0.039 & \cellcolor{KappaHalfBg}\textbf{0.914} \\

& PGN     
& \cellcolor{KappaStarker}107 & \cellcolor{KappaStarBg}0.032 & \cellcolor{KappaHalfBg}\underline{0.977}
& \cellcolor{KappaStarker}69  & \cellcolor{KappaStarBg}0.050 & \cellcolor{KappaHalfBg}\underline{0.999}
& \cellcolor{KappaStarker}85  & \cellcolor{KappaStarBg}0.041 & \cellcolor{KappaHalfBg}\underline{0.822}
& \cellcolor{KappaStarker}40  & \cellcolor{KappaStarBg}0.087 & \cellcolor{KappaHalfBg}\underline{0.890} \\

\midrule

{WideResNet-101} 
& DN      
& \cellcolor{KappaStarker}\textbf{51} & \cellcolor{KappaStarBg}\textbf{0.072} & \cellcolor{KappaHalfBg}0.973
& \cellcolor{KappaStarker}\textbf{32} & \cellcolor{KappaStarBg}\textbf{0.131} & \cellcolor{KappaHalfBg}0.987
& \cellcolor{KappaStarker}\textbf{37} & \cellcolor{KappaStarBg}\textbf{0.108} & \cellcolor{KappaHalfBg}0.915
& \cellcolor{KappaStarker}\textbf{23} & \cellcolor{KappaStarBg}\textbf{0.203} & \cellcolor{KappaHalfBg}0.947 \\

& FGSM    
& \cellcolor{KappaStarker}\underline{67} & \cellcolor{KappaStarBg}\underline{0.061} & \cellcolor{KappaHalfBg}0.871
& \cellcolor{KappaStarker}\underline{44} & \cellcolor{KappaStarBg}\underline{0.100} & \cellcolor{KappaHalfBg}0.838
& \cellcolor{KappaStarker}49 & \cellcolor{KappaStarBg}\underline{0.094} & \cellcolor{KappaHalfBg}0.732
& \cellcolor{KappaStarker}31 & \cellcolor{KappaStarBg}\underline{0.153} & \cellcolor{KappaHalfBg}0.822 \\

& MI-FGSM 
& \cellcolor{KappaStarker}\underline{67} & \cellcolor{KappaStarBg}0.049 & \cellcolor{KappaHalfBg}0.996
& \cellcolor{KappaStarker}48 & \cellcolor{KappaStarBg}0.080 & \cellcolor{KappaHalfBg}0.997
& \cellcolor{KappaStarker}\underline{42} & \cellcolor{KappaStarBg}0.093 & \cellcolor{KappaHalfBg}0.964
& \cellcolor{KappaStarker}\underline{29} & \cellcolor{KappaStarBg}0.152 & \cellcolor{KappaHalfBg}0.991 \\

& PGD-5   
& \cellcolor{KappaStarker}193 & \cellcolor{KappaStarBg}0.015 & \cellcolor{KappaHalfBg}\textbf{0.999}
& \cellcolor{KappaStarker}167 & \cellcolor{KappaStarBg}0.023 & \cellcolor{KappaHalfBg}\textbf{0.999}
& \cellcolor{KappaStarker}83  & \cellcolor{KappaStarBg}0.036 & \cellcolor{KappaHalfBg}0.920
& \cellcolor{KappaStarker}91  & \cellcolor{KappaStarBg}0.032 & \cellcolor{KappaHalfBg}0.940 \\

& PGD-20  
& \cellcolor{KappaStarker}296 & \cellcolor{KappaStarBg}0.011 & \cellcolor{KappaHalfBg}\textbf{0.999}
& \cellcolor{KappaStarker}202 & \cellcolor{KappaStarBg}0.020 & \cellcolor{KappaHalfBg}\textbf{0.999}
& \cellcolor{KappaStarker}79  & \cellcolor{KappaStarBg}0.032 & \cellcolor{KappaHalfBg}\textbf{0.997}
& \cellcolor{KappaStarker}66  & \cellcolor{KappaStarBg}0.045 & \cellcolor{KappaHalfBg}\textbf{0.997} \\

& PGN     
& \cellcolor{KappaStarker}83 & \cellcolor{KappaStarBg}0.037 & \cellcolor{KappaHalfBg}0.997
& \cellcolor{KappaStarker}65 & \cellcolor{KappaStarBg}0.050 & \cellcolor{KappaHalfBg}0.998
& \cellcolor{KappaStarker}59 & \cellcolor{KappaStarBg}0.048 & \cellcolor{KappaHalfBg}\underline{0.994}
& \cellcolor{KappaStarker}32 & \cellcolor{KappaStarBg}0.123 & \cellcolor{KappaHalfBg}\underline{0.995} \\
 \bottomrule
 \end{tabular}
 }
\end{table*}

\section{Experiments}
\label{sec:experiments}
In this section, we benchmark adversarial attacks using the proposed analysis across different architectures and settings, and study attack effectiveness, statistical behavior, and model confidence under perturbation risk evaluation. Finally, we present ablation studies of the proposed attack.

\subsection{Experimental setup}
\textbf{Attacks selected.}
In addition to the proposed DN attack, we consider a subset of common and widely used adversarial methods that well categorize distinct optimization behaviors. 
PGD \cite{pgd_attack} performs a direct steepest-ascent optimization to maximize the loss within the admissible perturbation set. While highly effective in worst-case settings, PGD and related strong attacks (e.g., CW \cite{Carlini017}, DF \cite{Moosavi16_deepfool}) primarily emphasize decision-boundary crossing rather than neighborhood exploration, less aligned with the objectives of this paper.
To explicitly account for stochastic optimization effects, 
we include MI-FGSM \cite{dong2018boosting} and PGN \cite{ge2023boosting}. MI-FGSM introduces momentum as a form of structured noise. PGN further extends this by injecting noise at each iteration and incorporating Hessian-based regularization.
Finally, we consider the simple FGSM \cite{Szegedy14}, which, despite its simplicity, exhibits informative behavior under our analysis.

Regarding attack settings, we consider both $\ell_\infty$ and $\ell_2$ threat models.
For attacks that use momentum, we adopt standard parameters with momentum $\mu = 1$, step size $\alpha = \varepsilon / T$, number of steps $T=10$, and $N = 10$ for both DN and PGN. For PGD, we use a step size $\alpha = 2/255$ for $\ell_\infty$ attacks and $\alpha = 0.25$ for $\ell_2$ attacks, and we evaluate it with both $5$ and 20 steps. Finally, for DN we used $\kappa_\text{adv} = 100$, and conduct ablation studies in Section \ref{sec:ablation}.

\textbf{Metrics.}
The directional perturbation risk $\mathcal{R}$
(Eq.~\ref{eq:dir_noisy_risk}) is estimated via a Monte Carlo approach using
$n = 256$ samples. Since $\mathcal{R}$ is defined at the sample level, we also report its average over the test set, denoted by $\bar{\mathcal{R}}$.
Additionally, we report two complementary indicators: $\kappa_{\bar{\mathcal{R}}=0.25}$ and $\bar{\mathcal{R}}_{\kappa^*}$, where $\kappa^* = D^{1/4}$.
The former can be interpreted as a plausibility risk score, defined as the smallest
concentration parameter $\kappa$ such that $\bar{\mathcal{R}} \ge 0.25$, with lower values indicating that the attack recognizes more statistically relevant failure regions.
The latter, $\bar{\mathcal{R}}_{\kappa^*}$, measures the perturbation risk at the reference
concentration $\kappa^*$ (see Sec.~\ref{sec:methodology} and  Sec.~\ref{sec:correlaton_kappa}), corresponding to success rate that remain statistically plausible under uniform noise on the $\ell_p$-sphere.

\textbf{Models and datasets.}
We run tests on 2{,}000 random samples from ImageNet-Val, using pretrained versions of ResNet-50~\cite{he2016deep_resnet}, WideResNet-101~\cite{Zagoruyko2016WideRN_wideresnet}, VGG-16~\cite{Simonyan2014VeryDC_vgg}, and ViT-B~\cite{dosovitskiy2020image}. Experiments on CIFAR-10 are reported in Appendix~\ref{app:cifar10_results}, and support the analysis in a lower-dimensional setting.
\begin{figure*}[h]
    \centering
    \begin{subfigure}{0.8\textwidth}
        \centering
        \includegraphics[width=0.8\textwidth]{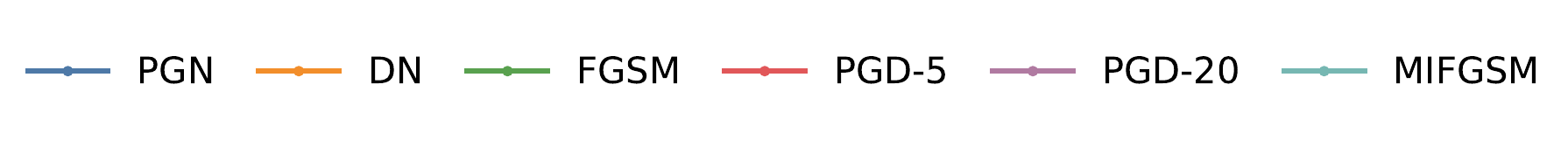}
    \end{subfigure}
    \begin{subfigure}{\textwidth}
    \begin{subfigure}{0.24\textwidth}
        \centering
        \includegraphics[width=\textwidth]{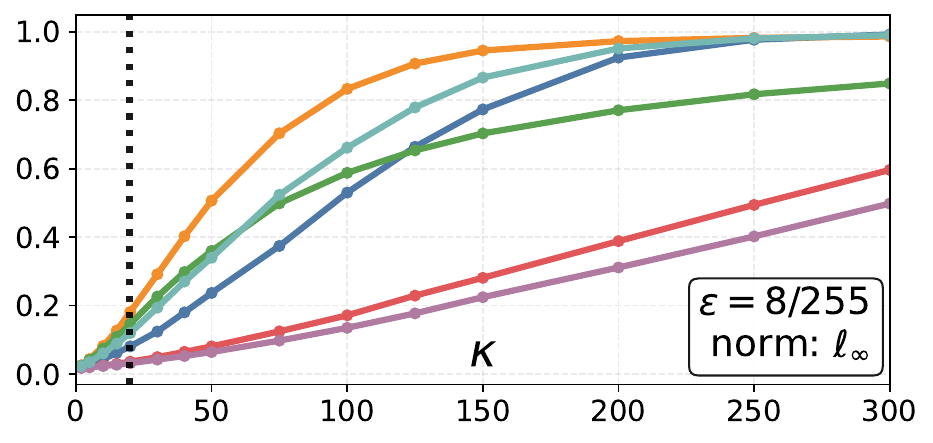}
    \end{subfigure}
    \begin{subfigure}{0.24\textwidth}
        \centering
        \includegraphics[width=\textwidth]{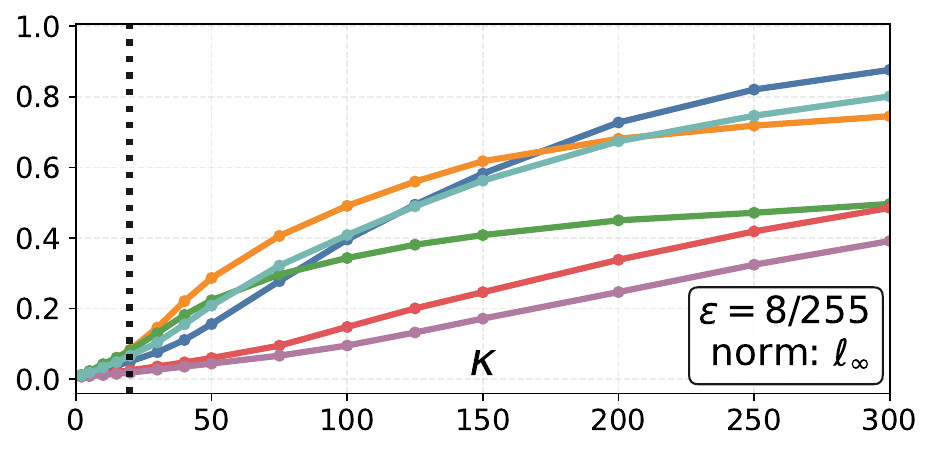}
    \end{subfigure}
    \begin{subfigure}{0.24\textwidth}
        \centering
        \includegraphics[width=\textwidth]{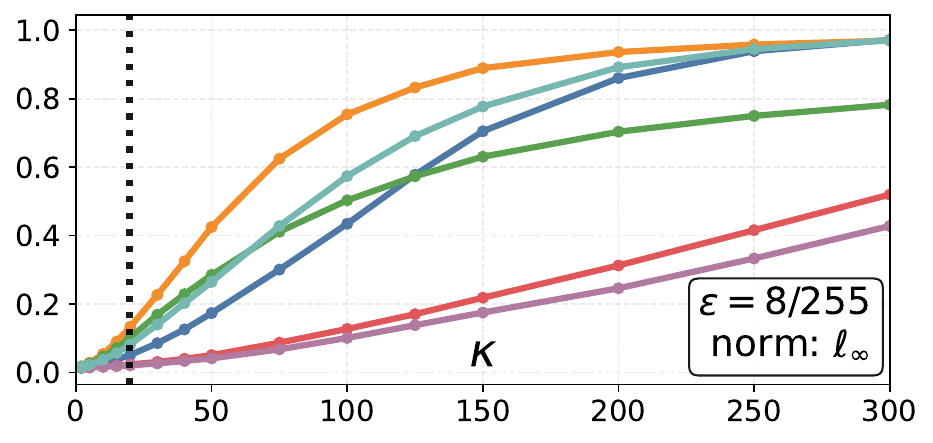}
    \end{subfigure}
    \begin{subfigure}{0.24\textwidth}
        \centering
        \includegraphics[width=\textwidth]{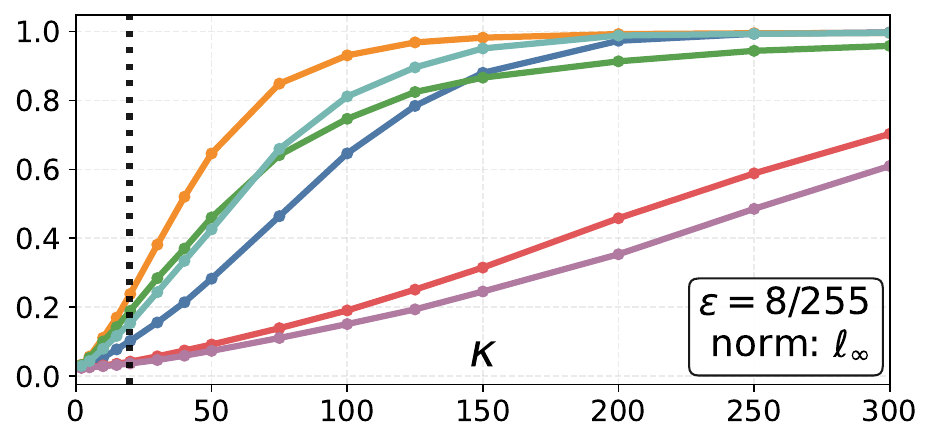}
    \end{subfigure}
    \end{subfigure}
\begin{subfigure}{\textwidth}
    \begin{subfigure}{0.24\textwidth}
        \centering
        \includegraphics[width=\textwidth]{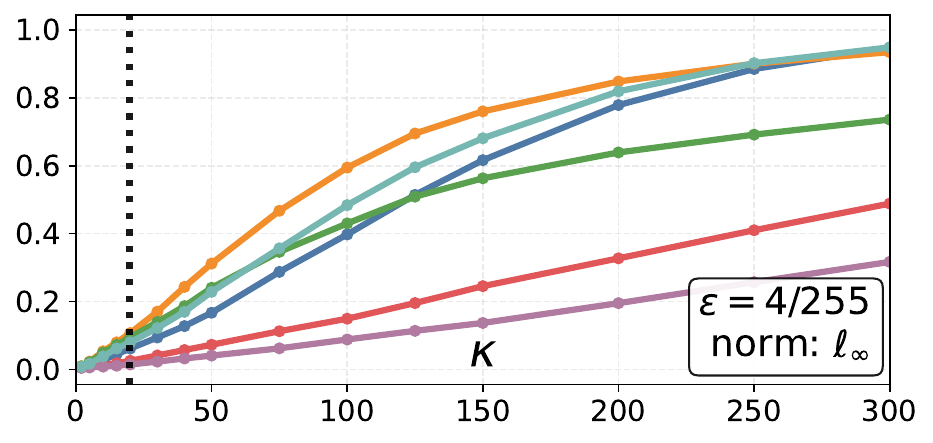}
    \end{subfigure}
    \begin{subfigure}{0.24\textwidth}
        \centering
        \includegraphics[width=\textwidth]{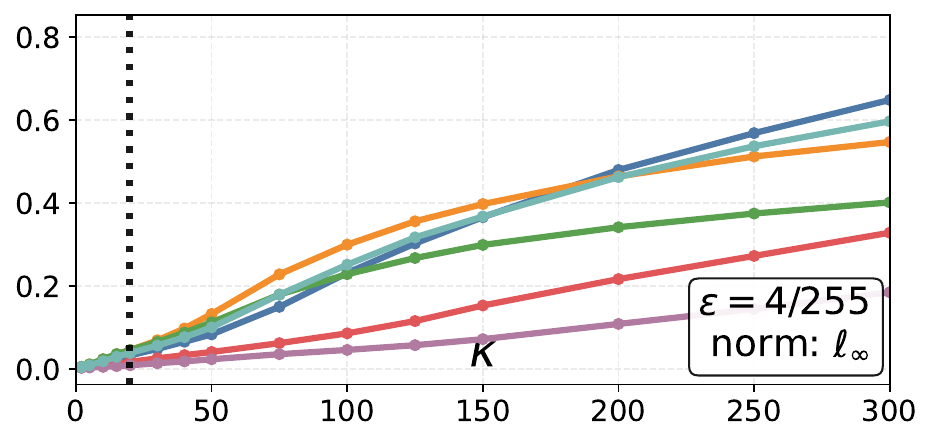}
    \end{subfigure}
    \begin{subfigure}{0.24\textwidth}
        \centering
        \includegraphics[width=\textwidth]{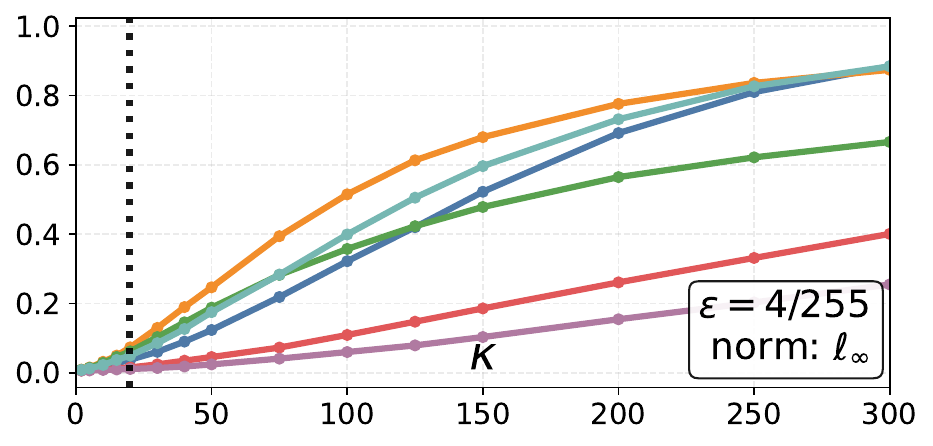}
    \end{subfigure}
    \begin{subfigure}{0.24\textwidth}
        \centering
        \includegraphics[width=\textwidth]{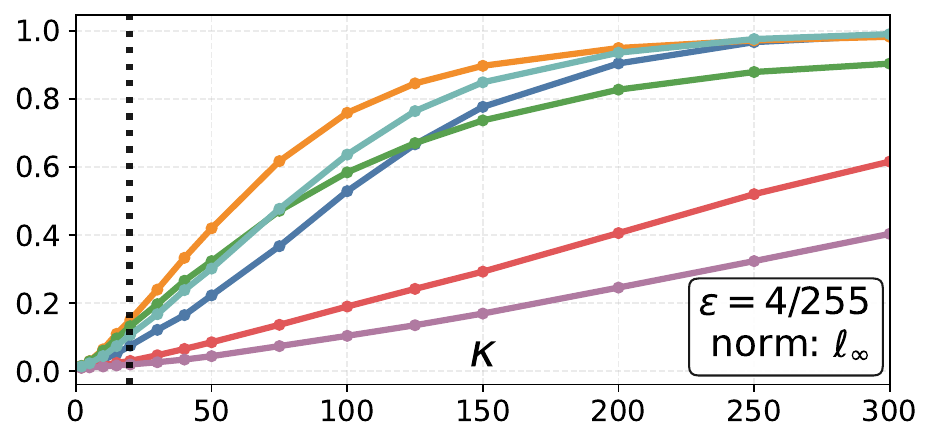}
    \end{subfigure}
    \end{subfigure}
    \begin{subfigure}{\textwidth}
    \begin{subfigure}{0.24\textwidth}
        \centering
        \includegraphics[width=\textwidth]{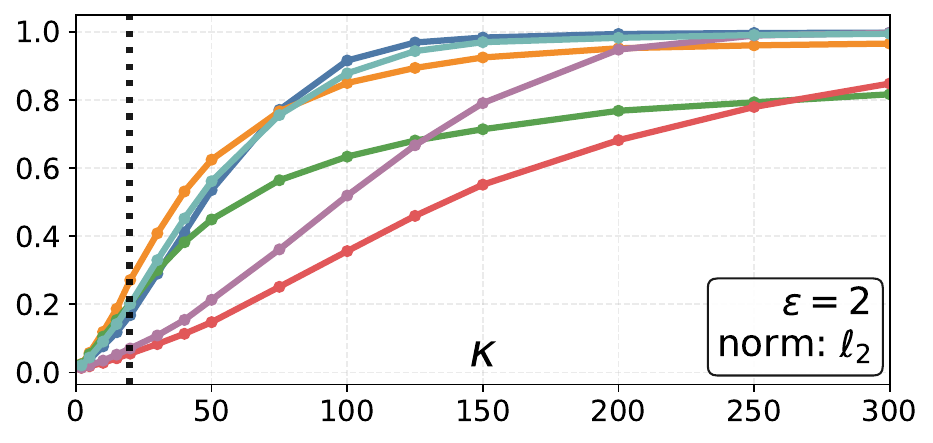}
    \end{subfigure}
    \begin{subfigure}{0.24\textwidth}
        \centering
        \includegraphics[width=\textwidth]{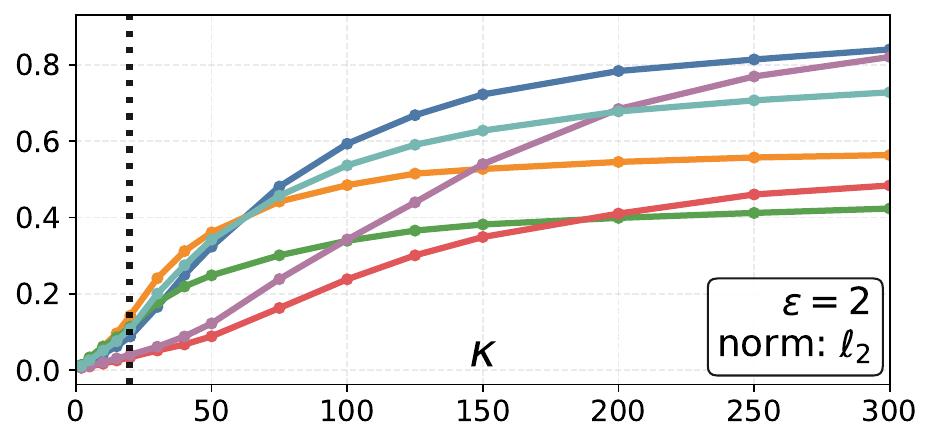}
    \end{subfigure}
    \begin{subfigure}{0.24\textwidth}
        \centering
        \includegraphics[width=\textwidth]{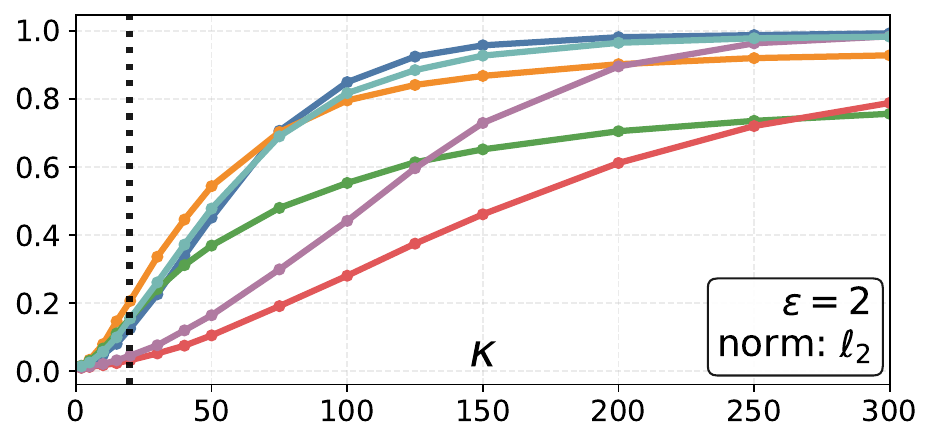}
    \end{subfigure}
    \begin{subfigure}{0.24\textwidth}
        \centering
        \includegraphics[width=\textwidth]{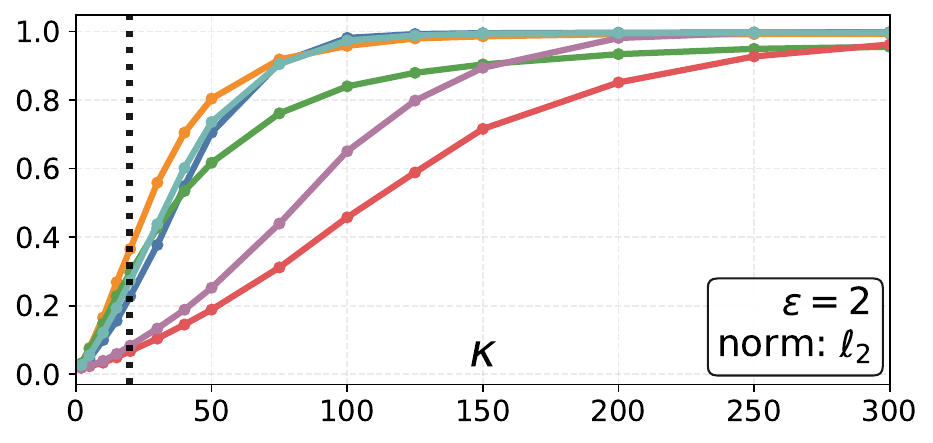}
    \end{subfigure}
    \end{subfigure}
\begin{subfigure}{\textwidth}
    \begin{subfigure}{0.24\textwidth}
        \centering
        \includegraphics[width=\textwidth]{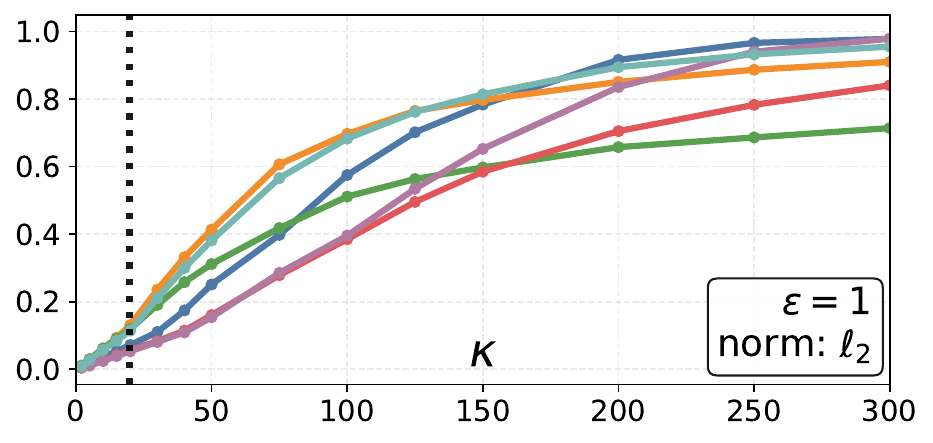}
        \caption{ResNet50}
    \end{subfigure}
    \begin{subfigure}{0.24\textwidth}
        \centering
        \includegraphics[width=\textwidth]{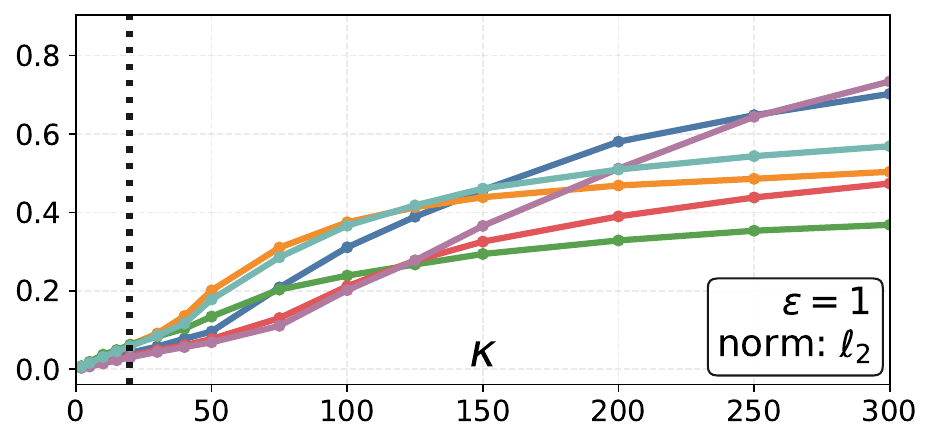}
        \caption{ViT-B}
    \end{subfigure}
    \begin{subfigure}{0.24\textwidth}
        \centering
        \includegraphics[width=\textwidth]{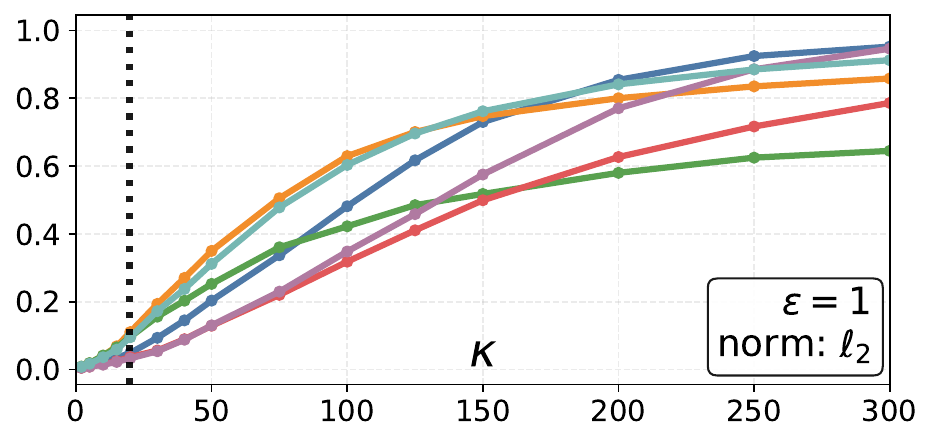}
        \caption{WideResNet101}
    \end{subfigure}
    \begin{subfigure}{0.24\textwidth}
        \centering
        \includegraphics[width=\textwidth]{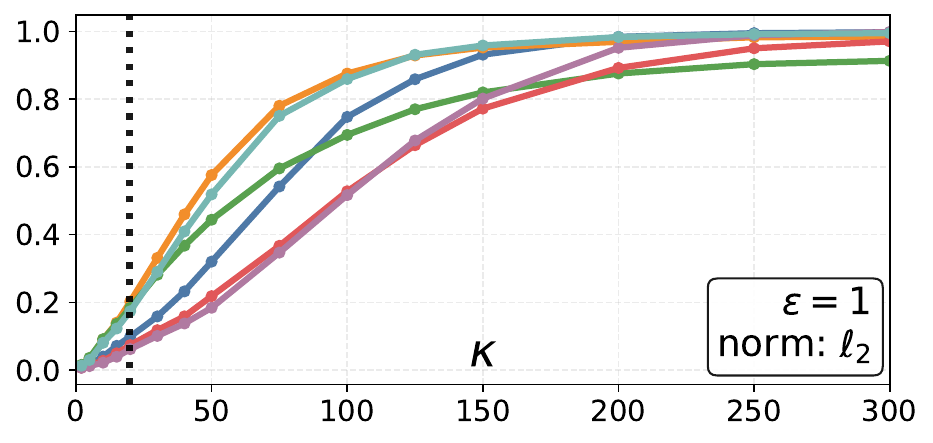}
        \caption{VGG16}
    \end{subfigure}
    \end{subfigure}
    \caption{\small{The average perturbation risk $\bar{\mathcal{R}}$ (y-axis) is shown as a function of $\kappa$ for different attacks under $\ell_\infty$ (first and second rows, $\varepsilon = 8/255$ and $4/255$) and $\ell_2$ (third and fourth, $\varepsilon = 0.5$ and $\varepsilon = 1.0$) threat models, evaluated on ResNet-50, ViT-B, WideResNet-101, and VGG-16. The reference value $\kappa^*$ is indicated by dotted lines, highlighting a statistically plausible risk regime.}}
    \label{f:example_attack}
\end{figure*}

\subsection{Benchmarking Adversarial Attacks}
\label{sec:benchmarks_attack}
To enable a meaningful comparison of adversarial attacks under the perturbation risk perspective, Table~\ref{table:imagenet_attacks} reports a benchmark of selected attacks based on the proposed metrics, while also including the classical attack success rate (ASR in table) for comparison. Results are reported for both $p=\infty$ and $p=2$ across different $\varepsilon$. Best and second-best are highlighted using boldface and underlining, respectively.

In all configurations, the proposed DN attack consistently achieves the highest directional perturbation risk in low regimes $\kappa$, as indicated by $\bar{\mathcal{R}}_{\kappa^*}$. FGSM often emerges as the second-best method at $ \kappa^* = D^{1/4}$, despite its simplicity and the fact that it is commonly regarded in the literature as a weak attack. Similar consideration also could be reported for the $\kappa_{\bar{\mathcal{R}}=0.25}$ indicator, showing the benefits of the above mentioned approaches for low $\kappa$ values. 
Conversely, this comparison highlights a key limitation of stronger techniques: attacks such as PGD, while highly effective in the pointwise limit (as shown by the ASR, i.e., $\bar{\mathcal{R}}$ for $\kappa \rightarrow \infty$), do not necessarily correspond to representative or stable failure modes when stochastic perturbations are taken into account. Consequently, ASR alone can be misleading, and even simple attacks such as FGSM may be more informative than stronger iterative ones when evaluated through a perturbation risk lens.
Interestingly, PGN exhibits limited perturbation risk: despite outperforming PGD in the proposed analysis, the use of a Hessian regularization, known to enhance transferability across architectures~\cite{ge2023boosting}, is effective mainly in high-$\kappa$ regimes. This questions the connection between stable loss curvature and robustness to statistically broad noisy perturbations.

\begin{figure}[h]
    \centering
    \begin{subfigure}{\columnwidth}
    \centering
    \begin{subfigure}{0.49\columnwidth}
        \centering
        \includegraphics[width=\columnwidth]{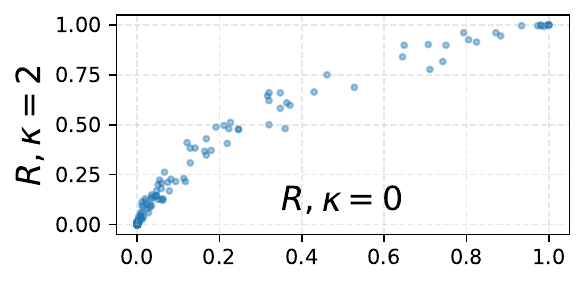}
    \end{subfigure}
    \begin{subfigure}{0.49\columnwidth}
    \centering
    \includegraphics[width=\columnwidth]{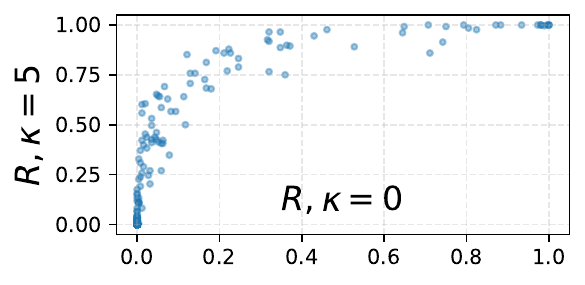}
    \end{subfigure}
    \begin{subfigure}{0.49\columnwidth}
    \centering
    \includegraphics[width=\columnwidth]{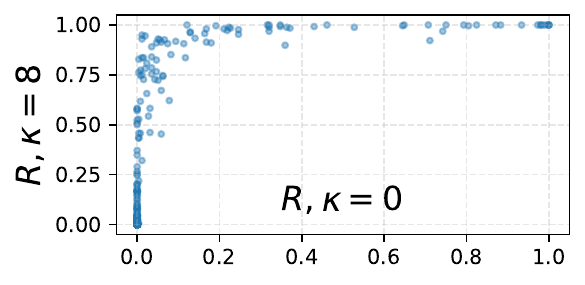}
    \end{subfigure}
    \begin{subfigure}{0.49\columnwidth}
    \centering
    \includegraphics[width=\columnwidth]{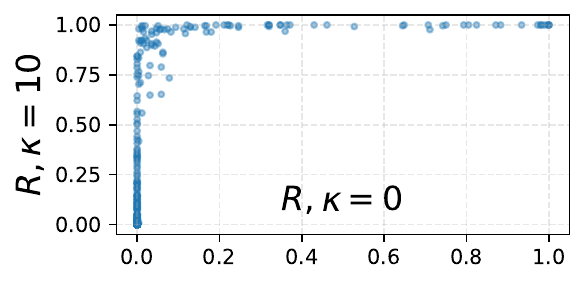}
    \end{subfigure}
    \caption{}
    \label{fig:corr_a}
    \end{subfigure}
    \begin{subfigure}{\columnwidth}
    \begin{subfigure}{\columnwidth}
    \centering
    \includegraphics[width=\columnwidth]{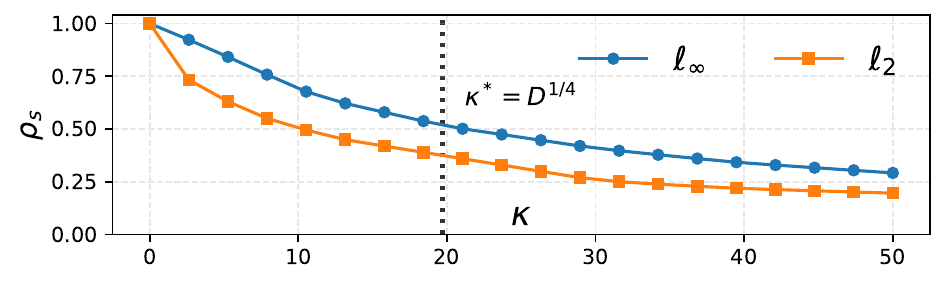}
    \end{subfigure}
    \begin{subfigure}{\columnwidth}
    \centering
    \includegraphics[width=\columnwidth]{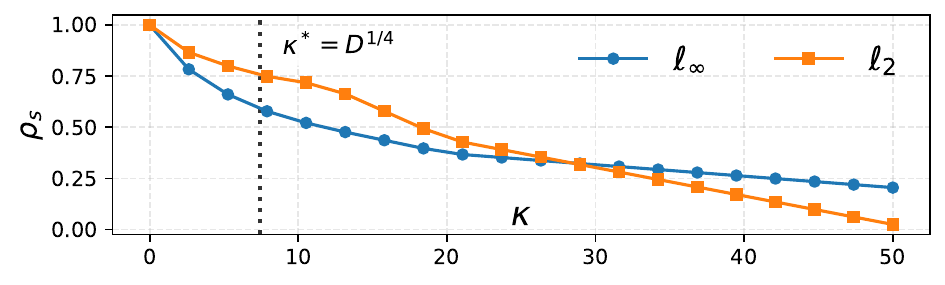}
    \end{subfigure}
    \caption{}
    \label{fig:corr_b}
    \end{subfigure}
    \caption{\small{Correlation between uniform project noise ($\mathcal{R}$ at $\kappa = 0$) and $\mathcal{R}$ varying $\kappa$, using DN.
\textbf{(a)} per-sample relation on ImageNet-Val samples for  $\ell_\infty$ with $\varepsilon = 16/255$.
\textbf{(b)} Spearman coefficient $\rho_s$ vs.\ $\kappa$ for ImageNet-Val (top: $\varepsilon = 16/255$ for $\ell_\infty$, $\varepsilon = 2.0$ for $\ell_2$) and CIFAR-10 (bottom: $\varepsilon = 8/255$ for $\ell_\infty$, $\varepsilon = 1.0$ for $\ell_2$)}}
    \label{f:correlation_analysis}
\end{figure}

\paragraph{Dynamics of $\bar{\mathcal{R}}$ with $\kappa$}
To further analyze attack behavior, Figure~\ref{f:example_attack} reports the evolution of $\bar{\mathcal{R}}$ as a function of the concentration parameter $\kappa$ up to $\kappa = 300$, corresponding to a high-concentration regime. A consistent trend emerges across all architectures for PGD, which increases $\bar{\mathcal{R}}$ only slowly as $\kappa$ grows, although it eventually achieves the highest attack success rates for values of $\kappa$ larger than $300$, as reported in Table~\ref{table:imagenet_attacks}.
As discussed in Section ~\ref{sec:methodology}, this behavior
indicates that PGD primarily exploits extremely narrow angular neighborhoods on the
$\ell_p$-sphere, highlighting the statistical rarity of such perturbations.

Importantly, the reference concentration $\kappa^* = D^{1/4}$ (vertical dotted lines)
reveals a substantial gap between $\bar{\mathcal{R}}$ and the ASR, even considering the DN attack.
From a probabilistic perspective, this gap highlights that standard adversarial attacks primarily exploit brittle failure modes, which have limited relevance under realistic perturbation distributions~\cite{fawzi2018analysis, carlini2019evaluating}.
Despite this common behavior, the DN attack exhibits a faster increase of $\bar{\mathcal{R}}$ with inital values of $\kappa$, indicating broader, noise-aligned failure regions at moderate concentrations. While not optimized for worst-case success, DN and FGSM remain more informative under perturbation risk evaluation, with similar, though more limited, trends observed for MIFGSM and PGN due to the use of noise during optimization.

\subsection{Empirical validation of $\kappa^*$}
\label{sec:correlaton_kappa}
The reference concentration $\kappa^* = D^{1/4}$ is introduced based on the analysis
presented in Section~\ref{sec:methodology} and Appendix~\ref{app:kappa_scaling}.
To further validate this choice from an empirical perspective, we analyze the behavior of
the directional perturbation risk for small values of $\kappa$, focusing on DN-based attacks, which, according to the previous benchmarks, achieve best performance in low $\kappa$ regimes.
Specifically, we study the sample-wise correlation between $\mathcal{R}$ computed at
fixed $\kappa$ values and that obtained under uniform projected noise ($\mathcal{R}$ with $\kappa = 0$).
As illustrated in Figure~\ref{fig:corr_a}, for small concentration values (e.g., $\kappa = 2$), the directional perturbation risk remains strongly correlated with the uniform-noise case; however, as $\kappa$ increases (e.g., $\kappa = 10$), this similarity progressively decreases, indicating that the attack achieves high success rates often in highly directional regimes, while remaining largely ineffective under uniform noise.

To quantify this trend across multiple samples, Figure~\ref{fig:corr_b} reports the Spearman rank correlation coefficient $\rho_s$ between the $\mathcal{R}$ values at a given $\kappa$ (x-axis) and those at $\kappa = 0$. Results are shown for both ImageNet-Val and CIFAR-10, and for $p=\infty$ and $p=2$, highlighting the effect of different input dimensionalities. 
In all settings, the rank correlation decreases sharply with $\kappa$, confirming that higher concentrations correspond to increasingly localized and less representative perturbation regimes.
The reference value $\kappa^* = D^{1/4}$ ($\approx 19$ for ImageNet, $\approx 7$ for CIFAR-10) lies in a transitional region where rank correlation with uniform noise remains relatively high, yet begins to decay rapidly. This regime represents a meaningful trade-off: it preserves a strong connection to noise-aligned perturbations while avoiding high-$\kappa$ regime where the analysis becomes dominated by corner-case and too worst-case behavior.

\begin{figure}[t]
    \centering
    \begin{subfigure}{0.49\columnwidth}
    \begin{subfigure}{\columnwidth}
        \centering
        \includegraphics[width=\textwidth]{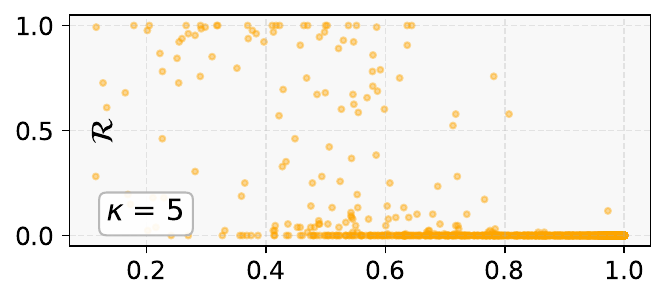}
    \end{subfigure}
    \begin{subfigure}{\columnwidth}
        \centering
        \includegraphics[width=\textwidth]{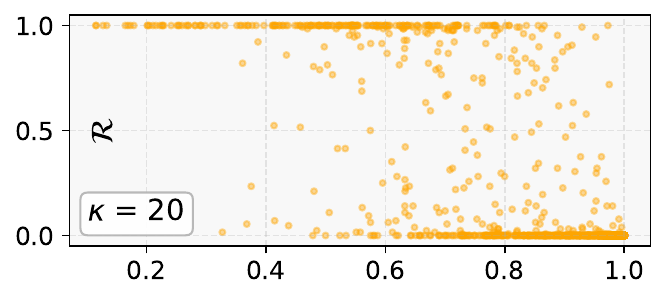}
    \end{subfigure}
    \begin{subfigure}{\columnwidth}
        \centering
        \includegraphics[width=\textwidth]{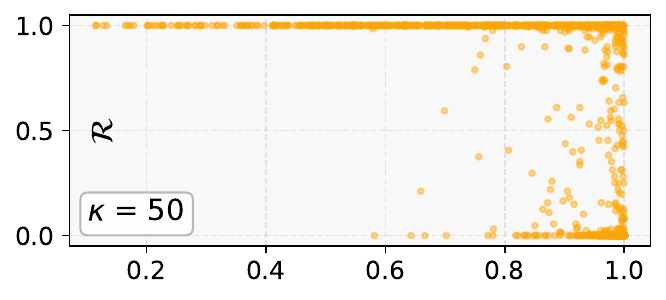}
    \end{subfigure}
    \caption{\small{ResNet50}}
    \label{fig:confidence_resnet50}
    \end{subfigure}
    \begin{subfigure}{0.49\columnwidth}
    \begin{subfigure}{\columnwidth}
        \centering
        \includegraphics[width=\textwidth]{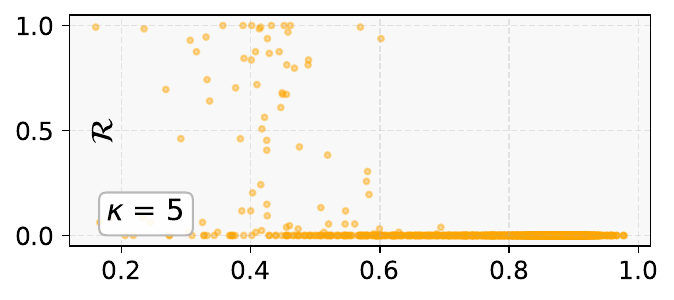}
    \end{subfigure}
    \begin{subfigure}{\columnwidth}
        \centering
        \includegraphics[width=\textwidth]{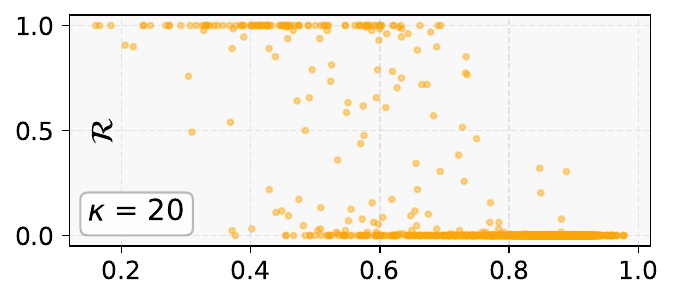}
    \end{subfigure}
    \begin{subfigure}{\columnwidth}
        \centering
        \includegraphics[width=\textwidth]{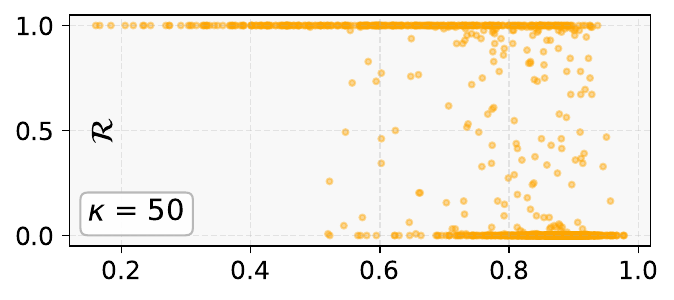}
    \end{subfigure}
    \caption{ViT-B}
    \label{fig:confidence_vit_b}
    \end{subfigure}
    \caption{\small{Analysis of $\mathcal{R}$ for ImageNet-Val samples, with $\varepsilon = 8/255$ and different values of $\kappa \in \{5, 20, 50\}$.
The x-axis reports the softmax probability of the predicted class.}}
    \label{fig:confidence_analysis}
\end{figure}

\subsection{On the reliability of model confidence}
\label{sec:confidence_analysis}
In addition to the analysis presented above, we investigate how model inference behavior relates to robustness across different $\kappa$ regimes, with the goal of identifying common characteristics of samples that are vulnerable to stochastic noise.
Figure~\ref{fig:confidence_analysis} reports the value of $\mathcal{R}$ for individual samples as a function of the model confidence on the corresponding clean input (x-axis), i.e., 
\(
\exp\big(f_y(x)\big)/
{\sum_{c} \exp\big(f_c(x)\big)}.
\)
The analysis is conducted for different concentration values, $\kappa \in {5, 20, 50}$, using the DN attack, illustrating the shift from noise-like to highly directional regimes.

For both ResNet50 and ViT-B, a consistent pattern emerges. In the lowest concentration regime ($\kappa = 5$), samples exhibiting high perturbation risk ($\mathcal{R} \approx 1$) are predominantly associated with low confidence scores. Conversely, samples predicted with high confidence remain largely robust, with $\mathcal{R}$ close to zero, despite being potentially vulnerable under classical worst-case adversarial evaluations.
As $\kappa$ increases to moderate values (e.g., $\kappa = 20 \approx \kappa^*$), high perturbation risk gradually extends toward samples with higher confidence. This effect is more pronounced for ResNet models, while ViT architectures exhibit more stable reliability, with high-risk samples remaining largely concentrated among low-confidence predictions. Only at relatively large values of $\kappa$ (e.g., $\kappa \geq 50$, where perturbation risk becomes less informative) does the distribution of highly attacked samples ($\mathcal{R}$ high) significantly include high-confidence predictions.

The trend suggests that, under statistically plausible perturbation regimes, adversarial vulnerability could be strongly coupled with model uncertainty. This provides a more nuanced perspective on the relationship between confidence and robustness: while confidence is known to be unreliable under worst-case threat models~\cite{grosse2018limitations, galil2021disrupting}, it could remain informative for identifying samples that are realistically at risk under stochastic perturbations. This observation highlights a potential connection between robustness evaluation and uncertainty estimation, suggesting that uncertainty-aware analyses~\cite{uncertaintygawlikowski2023survey, uncertaintyoberdiek2018classification} should not be dismissed based on their behavior under worst-case adversarial settings, and  still play a meaningful role in safety-oriented model assessment.
Results for additional models, exhibiting a similar trend, are reported in Appendix~\ref{app:confidence_extra}.

\subsection{Ablation studies on $\kappa_\text{adv}$}
\label{sec:ablation}
We conduct ablation studies to analyze the effect of the main parameters and settings of the DN attack. Across the evaluated configurations, we observe that increasing the values of $N$ and $T$ in Algorithm~\ref{alg:vmf_pgd} beyond $N > 10$ and $T > 10$ yields only marginal improvements, with slightly higher scores that do not justify the additional cost.

\begin{figure}[t]
    \centering
    \begin{subfigure}{\columnwidth}
        \centering
        \includegraphics[width=0.99\columnwidth]{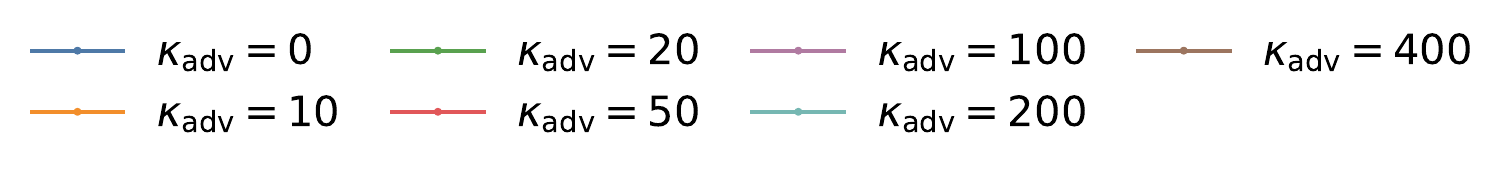}
    \end{subfigure}
    \begin{subfigure}{\columnwidth}
    \centering
    \includegraphics[width=0.99\columnwidth]{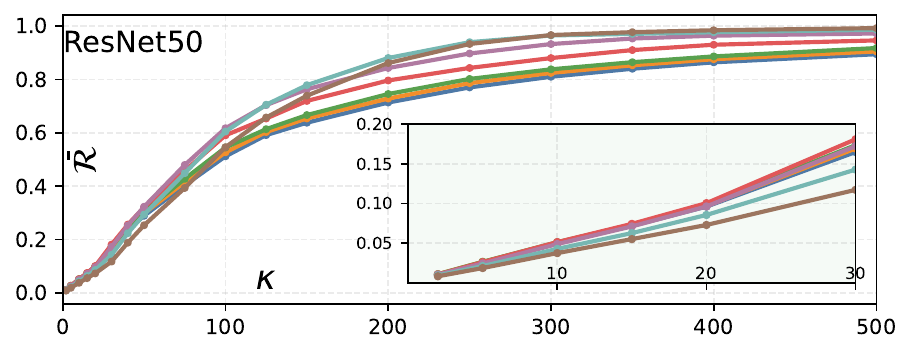}
    \end{subfigure}
    \begin{subfigure}{\columnwidth}
    \centering
    \includegraphics[width=0.99\columnwidth]{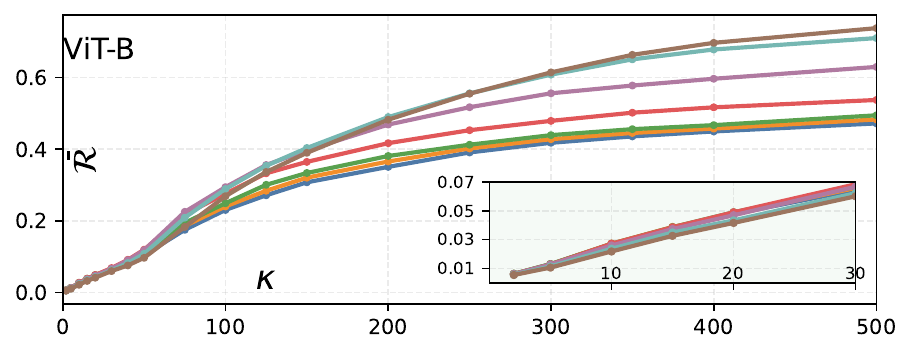}
    \end{subfigure}
    \caption{\small{Ablation study of the attack parameter $\kappa_{\textit{adv}}$. Plots show $\bar{\mathcal{R}}$ as $\kappa$ varies for different instances of the DN Attack, with $\kappa_{\textit{adv}}$ ranging from 0 to 200. Low-$\kappa$ regimes are shown in a zoomed-in view in the bottom-right corner.}}
    \label{f:ablation_attack}
\end{figure}

A more interesting parameter is concerns the choice of the reference concentration $\kappa_\text{adv}$ used during the attack optimization. Figure~\ref{f:ablation_attack} reports the behavior of $\mathcal{R}$ as a function of $\kappa$ on ImageNet-Val for different values of $\kappa_\text{adv}$. As expected, setting $\kappa_{\text{adv}} = 0$ leads to the weakest success with high $\kappa$, as it corresponds to optimizing over randomly sampled noise projected onto the $\varepsilon$-ball. In this case, adversarial effectiveness is mainly achieved at step $t = 0$ of Algorithm~\ref{alg:vmf_pgd}, where no noise is applied at the beginning.
As $\kappa_{\text{adv}}$ increases, the attack performs slightly worse in the low-$\kappa$ regimes (although the results are densely clustered for $\kappa_{\text{adv}} < 100$), but improves as $\kappa$ increases. Consistently, larger values of $\kappa_{\text{adv}}$ (e.g., $\kappa_{\text{adv}} = 200$) result in lower performance at small $\kappa$, as the attack focuses on extremely narrow neighborhoods, behaving similarly to a PGD-like optimization and becoming more effective only in high-$\kappa$ regimes.

As shown in Figure~\ref{f:ablation_attack}, we identify $\kappa_{\text{adv}} = 50$ and $\kappa_{\text{adv}} = 100$ as good trade-offs, achieving good success in low-$\kappa$ regimes (approximately comparable to that obtained with $\kappa_{\text{adv}} \approx \kappa^*$), while exhibiting better behavior at higher $\kappa$ values. Overall, we observe that fine-grained selection of $\kappa_{\text{adv}}$ leads to only marginal differences, whereas the choice of the value range is more important. As an empirical guideline, we recommend selecting $\kappa_{\text{adv}} \ll \sqrt{D}$ for effective analysis in low-$\kappa$ regimes, as further supported by the CIFAR-10 results reported in Appendix~\ref{app:cifar10_results}.
%

\section{Conclusion and Discussion}
\label{sec:conclusion}

This work examines the role of adversarial attacks in robustness and safety evaluation.
We introduce a probabilistic framework that bridges adversarial and noisy robustness through directional perturbation risk analysis, utilizing a concentration parameter to characterize the relationship between adversarial directions and noise-aligned perturbations.
We then presented a directional noisy attack that explicitly targets statistically representative failure regions. Results show that many standard adversarial attacks achieve high success only in highly worst-case regimes, while simpler or noise-aware strategies can be more informative when robustness is evaluated from a perturbation risk perspective.
These findings, rather than discouraging the use of attacks for this purpose, highlight important limitations of relying solely on attack success rates and invite practitioners to complement attack evaluation with perturbation risk analysis.

As future work, our results motivate further investigation into the reliability of model confidence as a proxy for statistical robustness (as suggested in Section~\ref{sec:confidence_analysis}). Moreover, extending the analysis beyond the specific $\varepsilon$-constrained setting is an important direction for safety assessment.
 

{\small
\section*{Acknowledgements}
This work was partially supported by project SERICS (PE00000014) under the MUR (Ministero dell'Università e della Ricerca) National Recovery and Resilience Plan funded by the European Union - NextGenerationEU.}

\bibliography{main}
\bibliographystyle{icml2026}

\newpage
\appendix
\onecolumn
\section*{\centering Appendix}
\addcontentsline{toc}{section}{Appendix}

\section{Sampling Directions on $\ell_p$-Spheres}
\label{app:lp_sampling}

Let $\xi \sim \mathcal{N}(0, I_d)$ be a standard Gaussian vector in $\mathbb{R}^d$.  
We consider perturbations obtained by projection onto the $\ell_p$-sphere of radius $r$,
\(
\eta = \Pi_{\|\cdot\|_p = r}(\xi).
\)

\paragraph{Exact uniformity for $p = 2$.}
When $p = 2$, the Gaussian distribution is isotropic and invariant under any orthogonal transformation $Q \in \mathrm{O}(d)$, i.e., $Q\xi \stackrel{d}{=} \xi$.  
Since $\|Q\xi\|_2 = \|\xi\|_2$, it follows that
\(
Q \frac{\xi}{\|\xi\|_2} \stackrel{d}{=} \frac{\xi}{\|\xi\|_2},
\)
implying that $\xi / \|\xi\|_2$ is invariant under all rotations.  
As a consequence, $\Pi_{\|\cdot\|_2 = r}(\xi)$ samples directions exactly uniformly on the $\ell_2$-sphere with respect to the surface-area (and cone) measure.

\paragraph{Approximate sampling for $p \neq 2$.}
For $p \neq 2$, the $\ell_p$ norm is not invariant under the full orthogonal group, but only under signed permutations of the coordinate axes.  
As a result, the projection $\xi / \|\xi\|_p$ does not preserve the rotational invariance of the Gaussian distribution, and the induced distribution on the $\ell_p$-sphere is not uniform with respect to either surface-area or cone measure.
Therefore, Gaussian projection yields an approximation rather than an exact uniform sampler for $p \neq 2$.

Despite this lack of exact uniformity, Gaussian projection remains a practical and effective approximation in high dimensions.  
Indeed, by concentration of measure,
\(
\|\xi\|_p^p = \sum_{i=1}^d |\xi_i|^p
\)
concentrates sharply around its expectation as $d$ increases. Consequently, normalization by $\|\xi\|_p$ primarily rescales $\xi$ by an almost deterministic factor, while preserving a diverse set of directions.
As a result, $\xi / \|\xi\|_p$ provides a stable set of noise-aligned perturbations that is sufficient for estimating perturbation risk probabilities, while retaining exact uniformity in the $\ell_2$ case.

As mentioned in Section \ref{sec:methodology}, the norm used to define noisy perturbations does not need to coincide with the norm used by the adversarial attack.
In particular, both the probabilistic analysis in Eq.~\ref{eq:dir_noisy_risk} and the optimization problem in Eq.~\ref{eq:optimization_problem} can be formulated using a norm $p_A$ for sampling noise directions (line~14 of Alg.\ref{alg:vmf_pgd}), while adversarial perturbations are constrained in a possibly different norm $p_B$ (line~20 of Alg.\ref{alg:vmf_pgd}).
For clarity and consistency, we adopt the same norm throughout the paper, but this choice does not affect the generality of the proposed framework.

\section{On the Scaling of $\kappa$ and $\kappa^*$}
\label{app:kappa_scaling}
In this section, we provide theoretical insight into the role of the concentration parameter $\kappa$ in
Eq.~\ref{eq:dir_noisy_risk} and motivate the reference choice $\kappa^\star = D^{1/4}$.
Let $\xi \sim \mathcal{N}(\kappa v, I)$, where $v \in \mathbb{S}^{D-1}$ is a fixed unit direction.
We decompose $\xi$ as
\(
\xi = \kappa v + z,
\), where $z \sim \mathcal{N}(0, I)$.
We are interested in how the directional bias induced by $\kappa v$ compares to the isotropic noise $z$
after projection onto the $\ell_2$-sphere.
Please note that, from also considerations in Appendix \ref{app:lp_sampling}, since projection preserves angular ordering in the $p=2$ case, and only mildly distorts it for other norms in high dimensions, the pre-projection geometry primarily governs the regime separation induced by $\kappa$.

By standard concentration results for Gaussian vectors,
\(
\|z\|_2^2 \sim \chi^2(D),
\)
and
\(
\|z\|_2 = \sqrt{D}\,(1 + o(1)),
\)
where $\chi^2(D)$ denotes a chi-squared random variable with $D$ degrees of freedom, and $o(1)$ indicates a term that vanishes as $D \to \infty$.
Hence, the magnitude of the isotropic component scales as $\sqrt{D}$, while the deterministic shift
has magnitude $\|\kappa v\|_2 = \kappa$.
This induces two asymptotic regimes:
(i) a \textit{noise-dominated regime} ($\kappa \ll \sqrt{D}$), where the contribution of $\kappa v$ is negligible
relative to $\|z\|_2$; and
(ii) a \textit{direction-dominated regime} ($\kappa \gtrsim \sqrt{D}$), where samples concentrate sharply around $v$,
approaching a worst-case adversarial direction.

To further characterize this separation, we study the angular alignment between $\xi$ and $v$,
quantified via the cosine similarity
\[
\cos(\theta)
=
\frac{\langle \xi, v \rangle}{\|\xi\|_2}
=
\frac{\kappa + \langle z, v \rangle}{\|\kappa v + z\|_2}.
\]
Since $\langle z, v \rangle \sim \mathcal{N}(0,1)$ and $\|\kappa v + z\|_2 \approx \sqrt{D + \kappa^2}$,
we obtain in expectation
\(
\mathbb{E}[\cos(\theta)]
\approx
\frac{\kappa}{\sqrt{D + \kappa^2}}.
\)

Thus, for $\kappa \ll \sqrt{D}$,
\(
\mathbb{E}[\cos(\theta)] \approx \frac{\kappa}{\sqrt{D}},
\)
indicating that directional alignment grows smoothly and remains small in the noise-dominated regime.
Choosing $\kappa^\star = D^{1/4}$ yields
\(
\mathbb{E}[\cos(\theta)] \approx D^{-1/4},
\)
which vanishes asymptotically while remaining strictly positive.
In contrast, when $\kappa \gtrsim \sqrt{D}$, $\mathbb{E}[\cos(\theta)] \to 1$, indicating that perturbations become
strongly aligned with $v$ and increasingly resemble worst-case adversarial directions.

This places $\kappa^\star$ well within the noise-dominated regime,
ensuring that samples remain broadly stochastic while introducing a controlled and measurable directional preference.
In typical high-dimensional vision settings ($D \approx 10^4$--$10^6$), this choice corresponds to moderate values of $\kappa$
that are empirically well below $\sqrt{D}$, yet sufficient to induce meaningful directional bias.
Please note that $\kappa^\star = D^{1/4}$ is not intended to be an optimal or universal choice.
Rather, it serves as a convenient reference point for the proposed analysis, lying well within the noise-dominated regime while still inducing a measurable directional bias.
The conclusions of this work do not rely on the exact value of $\kappa^\star$, but instead on the qualitative behavior of the perturbation risk metric across low-$\kappa$ regimes.
Similar trends are observed for a broad range of $\kappa$ values satisfying $\kappa \ll \sqrt{D}$, as further supported by the experimental analysis.

\begin{table*}[t]
\centering
\caption{\small{Robustness scores under $\ell_\infty$ attacks ($\varepsilon \in \{4/255, 8/255\}$) and $\ell_2$ attacks ($\varepsilon \in \{0.5, 1.0\}$). 
Here, $\bar{\mathcal{R}}_{\kappa^*}$ denotes the average perturbation risk on the test set evaluated at $\kappa^* = D^{1/4}$, while $\kappa_{0.25}$ denotes the smallest concentration parameter $\kappa$ for which $\bar{\mathcal{R}} \ge 0.25$. 
ASR denotes the standard attack success rate, reported for comparison.
}}
\label{table:cifar10_results}
\resizebox{0.95\textwidth}{!}{
\begin{tabular}{ll|c|c|c||c|c|c||c|c|c||c|c|c}
\toprule
\cmidrule(lr){3-8}\cmidrule(lr){9-14}
& & \multicolumn{3}{c||}{$\ell_\infty$,  $\varepsilon = 4/255$} & \multicolumn{3}{c||}{$\ell_\infty$,  $\varepsilon = 8/255$}
& \multicolumn{3}{c||}{$\ell_2$,  $\varepsilon = 0.5$} & \multicolumn{3}{c}{$\ell_2$, $\varepsilon = 1.0$} \\
Model & Attack
& $k_{\bar{\mathcal{R}}={0.25}}\!\downarrow$ & $\bar{\mathcal{R}}_{\kappa^*}\!\uparrow$ & $\textit{ASR}\!\uparrow$
& $k_{\bar{\mathcal{R}}={0.25}}\!\downarrow$ & $\bar{\mathcal{R}}_{\kappa^*}\!\uparrow$  & $\textit{ASR}\!\uparrow$
& $k_{\bar{\mathcal{R}}={0.25}}\!\downarrow$ & $\bar{\mathcal{R}}_{\kappa^*}\!\uparrow$  & $\textit{ASR}\!\uparrow$
& $k_{\bar{\mathcal{R}}={0.25}}\!\downarrow$ & $\bar{\mathcal{R}}_{\kappa^*}\!\uparrow$  & $\textit{ASR}\!\uparrow$ \\
\midrule
{ResNet-32} 
& DN - ${k}_\text{adv} = 10$    
& \cellcolor{KappaStarker}{23}  & \cellcolor{KappaStarBg}\textbf{0.067} & \cellcolor{KappaHalfBg}0.81
& \cellcolor{KappaStarker}\underline{12}  & \cellcolor{KappaStarBg}\textbf{0.13} & \cellcolor{KappaHalfBg}0.93
& \cellcolor{KappaStarker}\textbf{7} & \cellcolor{KappaStarBg}\textbf{0.25} & \cellcolor{KappaHalfBg}0.874
& \cellcolor{KappaStarker}\textbf{4} & \cellcolor{KappaStarBg}\textbf{0.43} &  \cellcolor{KappaHalfBg}0.95 \\
& DN - ${k}_\text{adv} = 20$     
& \cellcolor{KappaStarker}\textbf{21} & \cellcolor{KappaStarBg}\textbf{0.067} & \cellcolor{KappaHalfBg}{0.86}
& \cellcolor{KappaStarker}\textbf{11} & \cellcolor{KappaStarBg}\textbf{0.13} & \cellcolor{KappaHalfBg}{0.97}
& \cellcolor{KappaStarker}\underline{8} & \cellcolor{KappaStarBg}\underline{0.22} & \cellcolor{KappaHalfBg}0.947
& \cellcolor{KappaStarker}\underline{5} & \cellcolor{KappaStarBg}\underline{0.40} & \cellcolor{KappaHalfBg}0.989 \\
& DN - ${k}_\text{adv} = 50$    
& \cellcolor{KappaStarker}\textbf{21}  & \cellcolor{KappaStarBg}\underline{0.061} & \cellcolor{KappaHalfBg}0.948
& \cellcolor{KappaStarker}{13}  & \cellcolor{KappaStarBg}\underline{0.11} & \cellcolor{KappaHalfBg}0.996
& \cellcolor{KappaStarker}{11} & \cellcolor{KappaStarBg}{0.13} & \cellcolor{KappaHalfBg}0.992
& \cellcolor{KappaStarker}{6} & \cellcolor{KappaStarBg}{0.278} &  \cellcolor{KappaHalfBg}\underline{0.999} \\
& DN - ${k}_\text{adv} = 100$     
& \cellcolor{KappaStarker}\underline{22} & \cellcolor{KappaStarBg}{0.056} & \cellcolor{KappaHalfBg}{0.978}
& \cellcolor{KappaStarker}{13} & \cellcolor{KappaStarBg}{0.097} & \cellcolor{KappaHalfBg}{0.999}
& \cellcolor{KappaStarker}{22} & \cellcolor{KappaStarBg}{0.11} & \cellcolor{KappaHalfBg}\underline{0.997}
& \cellcolor{KappaStarker}{7} & \cellcolor{KappaStarBg}{0.252} & \cellcolor{KappaHalfBg}\underline{0.999} \\
\cdashline{2-14}
& FGSM    
& \cellcolor{KappaStarker}31 & \cellcolor{KappaStarBg}0.057 & \cellcolor{KappaHalfBg}0.656
& \cellcolor{KappaStarker}17  & \cellcolor{KappaStarBg}0.10 & \cellcolor{KappaHalfBg}0.737
& \cellcolor{KappaStarker}12 & \cellcolor{KappaStarBg}0.16 & \cellcolor{KappaHalfBg}0.564
& \cellcolor{KappaStarker}8 & \cellcolor{KappaStarBg}0.231 & \cellcolor{KappaHalfBg}0.634 \\
& MI-FGSM 
& \cellcolor{KappaStarker}\underline{23}  & \cellcolor{KappaStarBg}{0.056} & \cellcolor{KappaHalfBg}0.98
& \cellcolor{KappaStarker}\underline{14}  & \cellcolor{KappaStarBg}{0.103} & \cellcolor{KappaHalfBg}0.997
& \cellcolor{KappaStarker}8 & \cellcolor{KappaStarBg}0.20 & \cellcolor{KappaHalfBg}0.943
& \cellcolor{KappaStarker}6 & \cellcolor{KappaStarBg}0.311 & \cellcolor{KappaHalfBg}0.997 \\
& PGD-5   
& \cellcolor{KappaStarker}52 & \cellcolor{KappaStarBg}0.022 & \cellcolor{KappaHalfBg}0.994
& \cellcolor{KappaStarker}40 & \cellcolor{KappaStarBg}0.028 & \cellcolor{KappaHalfBg}0.995
& \cellcolor{KappaStarker}15 & \cellcolor{KappaStarBg}0.088 & \cellcolor{KappaHalfBg}0.974
& \cellcolor{KappaStarker}15 & \cellcolor{KappaStarBg}0.111 & \cellcolor{KappaHalfBg}{0.95} \\
& PGD-20  
& \cellcolor{KappaStarker}61 & \cellcolor{KappaStarBg}0.018 & \cellcolor{KappaHalfBg}\textbf{0.999}
& \cellcolor{KappaStarker}44 & \cellcolor{KappaStarBg}0.032 & \cellcolor{KappaHalfBg}\textbf{1.000}
& \cellcolor{KappaStarker}18 & \cellcolor{KappaStarBg}0.063 & \cellcolor{KappaHalfBg}\textbf{0.999}
& \cellcolor{KappaStarker}14 & \cellcolor{KappaStarBg}0.108 & \cellcolor{KappaHalfBg}\textbf{1.000} \\
& PGN     
& \cellcolor{KappaStarker}24 & \cellcolor{KappaStarBg}0.048 & \cellcolor{KappaHalfBg}\underline{0.991}
& \cellcolor{KappaStarker}17  & \cellcolor{KappaStarBg}0.079 & \cellcolor{KappaHalfBg}\textbf{0.999}
& \cellcolor{KappaStarker}15 & \cellcolor{KappaStarBg}0.086 & \cellcolor{KappaHalfBg}0.996
& \cellcolor{KappaStarker}8 & \cellcolor{KappaStarBg}0.214 & \cellcolor{KappaHalfBg}\textbf{1.000} \\
\midrule
{VGG-16} 
%
%
& DN -- ${k}_{\text{adv}} = 10$
& \cellcolor{KappaStarker}44 & \cellcolor{KappaStarBg}\textbf{0.035} & \cellcolor{KappaHalfBg}0.585
& \cellcolor{KappaStarker}20 & \cellcolor{KappaStarBg}\underline{0.076} & \cellcolor{KappaHalfBg}0.728
& \cellcolor{KappaStarker}\underline{12} & \cellcolor{KappaStarBg}\textbf{0.142} & \cellcolor{KappaHalfBg}0.629
& \cellcolor{KappaStarker}\textbf{6}  & \cellcolor{KappaStarBg}\textbf{0.289} & \cellcolor{KappaHalfBg}0.745 \\
& DN -- ${k}_{\text{adv}} = 20$
& \cellcolor{KappaStarker}\underline{40} & \cellcolor{KappaStarBg}\textbf{0.035} & \cellcolor{KappaHalfBg}0.621
& \cellcolor{KappaStarker}18 & \cellcolor{KappaStarBg}\textbf{0.077} & \cellcolor{KappaHalfBg}0.775
& \cellcolor{KappaStarker}\textbf{11} & \cellcolor{KappaStarBg}\underline{0.132} & \cellcolor{KappaHalfBg}0.708
& \cellcolor{KappaStarker}\underline{7}  & \cellcolor{KappaStarBg}\underline{0.279} & \cellcolor{KappaHalfBg}0.818 \\
& DN -- ${k}_{\text{adv}} = 50$
& \cellcolor{KappaStarker}\textbf{37} & \cellcolor{KappaStarBg}\underline{0.033} & \cellcolor{KappaHalfBg}0.691
& \cellcolor{KappaStarker}19 & \cellcolor{KappaStarBg}0.064 & \cellcolor{KappaHalfBg}0.861
& \cellcolor{KappaStarker}15 & \cellcolor{KappaStarBg}0.081 & \cellcolor{KappaHalfBg}0.817
& \cellcolor{KappaStarker}9  & \cellcolor{KappaStarBg}0.196 & \cellcolor{KappaHalfBg}0.921 \\
& DN -- ${k}_{\text{adv}} = 100$
& \cellcolor{KappaStarker}\textbf{37} & \cellcolor{KappaStarBg}0.032 & \cellcolor{KappaHalfBg}0.726
& \cellcolor{KappaStarker}20 & \cellcolor{KappaStarBg}0.057 & \cellcolor{KappaHalfBg}0.899
& \cellcolor{KappaStarker}17 & \cellcolor{KappaStarBg}0.067 & \cellcolor{KappaHalfBg}0.866
& \cellcolor{KappaStarker}10 & \cellcolor{KappaStarBg}0.171 & \cellcolor{KappaHalfBg}0.960 \\
\cdashline{2-14}
%
& FGSM
& \cellcolor{KappaStarker}60 & \cellcolor{KappaStarBg}0.031 & \cellcolor{KappaHalfBg}0.513
& \cellcolor{KappaStarker}28 & \cellcolor{KappaStarBg}0.064 & \cellcolor{KappaHalfBg}0.599
& \cellcolor{KappaStarker}17 & \cellcolor{KappaStarBg}0.113 & \cellcolor{KappaHalfBg}0.477
& \cellcolor{KappaStarker}10  & \cellcolor{KappaStarBg}0.194 & \cellcolor{KappaHalfBg}0.545 \\
& MI-FGSM
& \cellcolor{KappaStarker}38 & \cellcolor{KappaStarBg}0.031 & \cellcolor{KappaHalfBg}0.739
& \cellcolor{KappaStarker}21 & \cellcolor{KappaStarBg}0.057 & \cellcolor{KappaHalfBg}0.909
& \cellcolor{KappaStarker}12 & \cellcolor{KappaStarBg}0.126 & \cellcolor{KappaHalfBg}0.693
& \cellcolor{KappaStarker}8  & \cellcolor{KappaStarBg}0.225 & \cellcolor{KappaHalfBg}0.860 \\
& PGD-5
& \cellcolor{KappaStarker}76 & \cellcolor{KappaStarBg}0.014 & \cellcolor{KappaHalfBg}\underline{0.876}
& \cellcolor{KappaStarker}52 & \cellcolor{KappaStarBg}0.021 & \cellcolor{KappaHalfBg}\underline{0.955}
& \cellcolor{KappaStarker}20 & \cellcolor{KappaStarBg}0.056 & \cellcolor{KappaHalfBg}0.809
& \cellcolor{KappaStarker}18 & \cellcolor{KappaStarBg}0.074 & \cellcolor{KappaHalfBg}0.894 \\
& PGD-20
& \cellcolor{KappaStarker}84 & \cellcolor{KappaStarBg}0.012 & \cellcolor{KappaHalfBg} \textbf{0.974}
& \cellcolor{KappaStarker}56 & \cellcolor{KappaStarBg}0.018 & \cellcolor{KappaHalfBg}\textbf{0.999}
& \cellcolor{KappaStarker}23 & \cellcolor{KappaStarBg}0.041 & \cellcolor{KappaHalfBg}\textbf{0.966}
& \cellcolor{KappaStarker}17 & \cellcolor{KappaStarBg}0.078 & \cellcolor{KappaHalfBg}\textbf{0.999} \\
& PGN
& \cellcolor{KappaStarker}39 & \cellcolor{KappaStarBg}0.029 & \cellcolor{KappaHalfBg}0.788
& \cellcolor{KappaStarker}23 & \cellcolor{KappaStarBg}0.050 & \cellcolor{KappaHalfBg}\underline{0.955}
& \cellcolor{KappaStarker}20 & \cellcolor{KappaStarBg}0.054 & \cellcolor{KappaHalfBg}\underline{0.922}
& \cellcolor{KappaStarker}10 & \cellcolor{KappaStarBg}0.147 & \cellcolor{KappaHalfBg}\underline{0.986} \\
 \bottomrule
 \end{tabular}
 }
\end{table*}

\begin{figure}[t]
    \centering
    \begin{subfigure}{\columnwidth}
    \centering
        \includegraphics[width=0.7\textwidth]{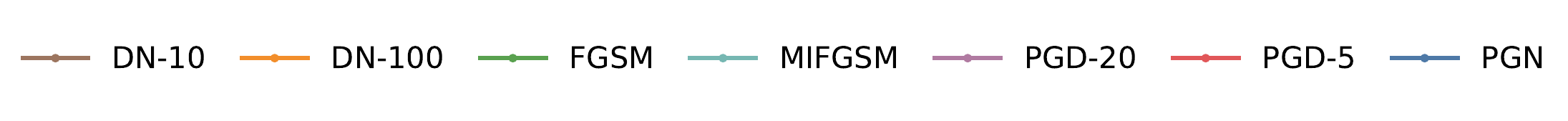}
    \end{subfigure}
    \begin{subfigure}{0.49\columnwidth}
    \begin{subfigure}{0.49\textwidth}
        \centering
        \includegraphics[width=\textwidth]{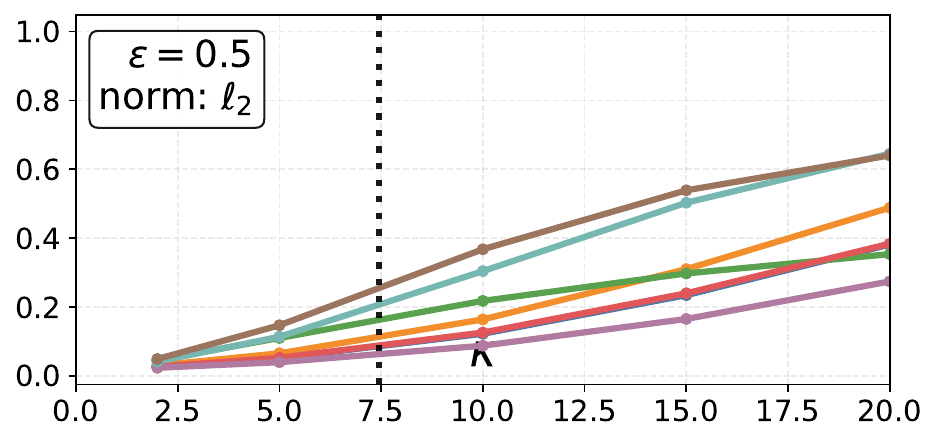}
    \end{subfigure}
    \begin{subfigure}{0.49\textwidth}
        \centering
        \includegraphics[width=\textwidth]{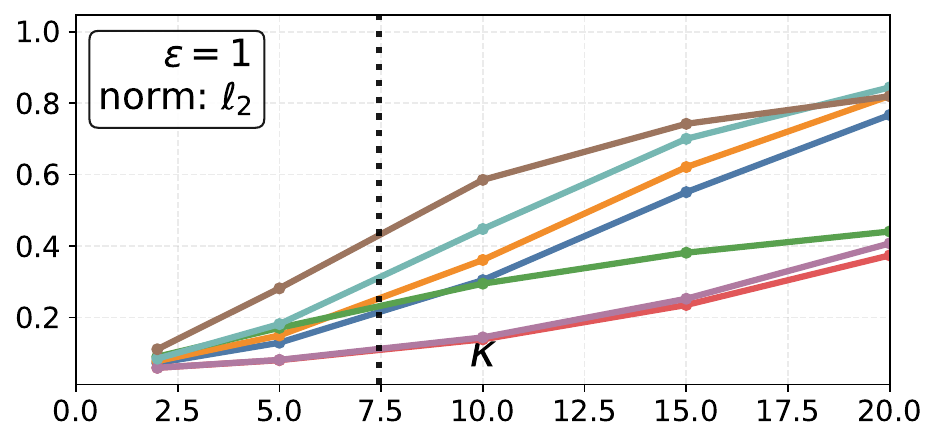}
    \end{subfigure}
    \caption{\small{ResNet32}}
    \end{subfigure}
    \begin{subfigure}{0.49\columnwidth}
    \begin{subfigure}{0.49\textwidth}
        \centering
        \includegraphics[width=\textwidth]{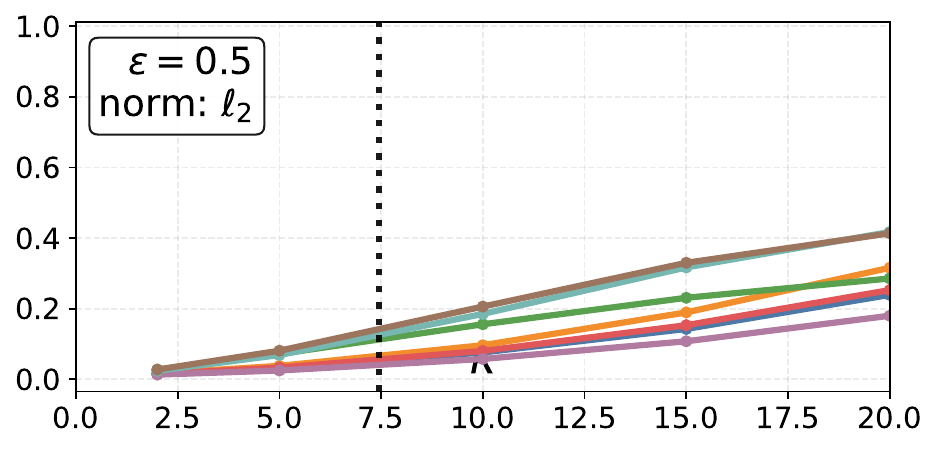}
    \end{subfigure}
    \begin{subfigure}{0.49\textwidth}
        \centering
        \includegraphics[width=\textwidth]{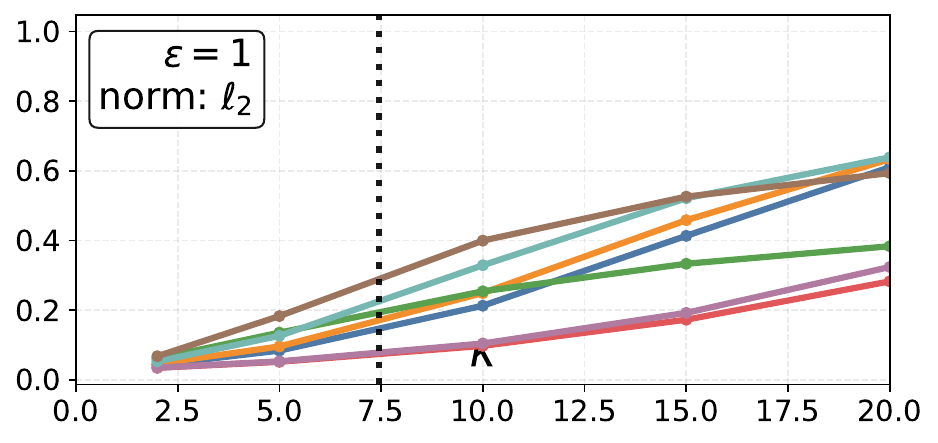}
    \end{subfigure}
    \caption{\small{VGG16}}
    \end{subfigure}
    \caption{\small{Comparison of different adversarial attacks on the CIFAR10 tests set measured by the perturbation risk metric $\bar{\mathcal{R}}$ at small values of $\kappa$ for ResNet-32 and VGG-16 under $\ell_2$ perturbations. The dotted lines denote the reference point $\kappa^*= D^{1/4}$.}}
    \label{fig:cifar10_curves}
\end{figure}

\section{Results on CIFAR-10}
\label{app:cifar10_results}
Table~\ref{table:cifar10_results} reports experimental results on the CIFAR-10 test set.
In this setting, the input dimensionality is $D = 32 \times 32 \times 3$, yielding a reference value $\kappa^* = D^{1/4} \approx 7$.
We consider CIFAR-10 pretrained versions of ResNet32 and VGG16\footnote{We use pretrained models available at \url{https://github.com/huyvnphan/PyTorch_CIFAR10}.}.

Following the same evaluation protocol adopted for ImageNet in Section~\ref{sec:experiments}, we analyze the directional perturbation risk metric at $\kappa = \kappa^*$, denoted by $\bar{\mathcal{R}}_{\kappa^*}$, as well as the smallest value of $\kappa$ for which $\bar{\mathcal{R}} = 0.25$.
The table also includes an explicit ablation over the adversarial concentration parameter $\kappa_{\text{adv}}$, obtained by considering different variants of the DN attack.
In this case, $\kappa_{\text{adv}}$ is explored over a lower range compared to Section~\ref{sec:ablation}, reflecting the reduced input dimensionality of CIFAR-10.
As a consequence, the suboptimal or transition values of $\kappa_{\text{adv}}$ differ from those observed in  Table~\ref{table:imagenet_attacks}, consistently with the dimensional scaling discussed in Appendix~\ref{app:kappa_scaling}.

Overall, the CIFAR-10 results further support the proposed analysis and confirm that the directional perturbation risk provides a more informative characterization of attack effectiveness in low-$\kappa$ regimes.
In particular, when $\kappa_{\text{adv}}$ is close to $\kappa^*$, the metric highlights whether an adversarial perturbation corresponds to a statistically representative noisy failure region, rather
than an isolated worst-case direction.
Consistent with the ImageNet analysis, we also observe a clear trade-off between adversarial success rate and robustness under stochastic perturbations.
While strong iterative attacks such as PGD-5 and PGD-20 achieve high adversarial success rates, they often exhibit limited effectiveness in low-$\kappa$ regimes, indicating that their success is concentrated along highly localized directions.
In contrast, attacks that maintain higher perturbation risk at low $\kappa$ correspond to broader and more stable failure regions, which are more relevant from a safety-oriented perspective.

We also report in Figure ~\ref{fig:cifar10_curves} the analysis of $\bar{\mathcal{R}}$ for attacks with $p=2$ as $\kappa$ varies, focusing on small $\kappa$ values that are closer to the informative noise regime (see Section~\ref{sec:correlaton_kappa}).
Consistent with the ImageNet analysis, the use of DN attacks exhibits a steeper increase in success rate in these low-$\kappa$ regimes.
At the same time, for all attacks, the curves clearly highlight the gap with respect to the classical adversarial success rate observed in the limit $\kappa \to \infty$, as reported in Table~\ref{table:cifar10_results}.

\begin{figure}[h]
    \centering
    \begin{subfigure}{0.8\columnwidth}
    \begin{subfigure}{0.33\columnwidth}
        \centering
        \includegraphics[width=\textwidth]{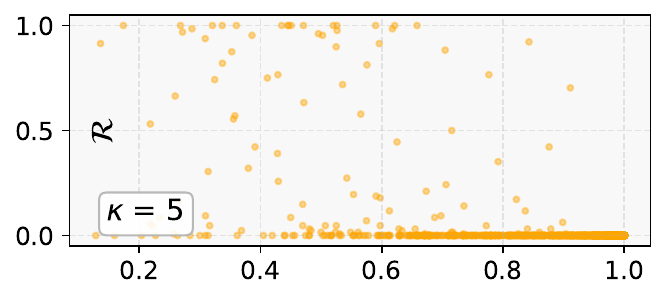}
    \end{subfigure}
    \begin{subfigure}{0.33\columnwidth}
        \centering
        \includegraphics[width=\textwidth]{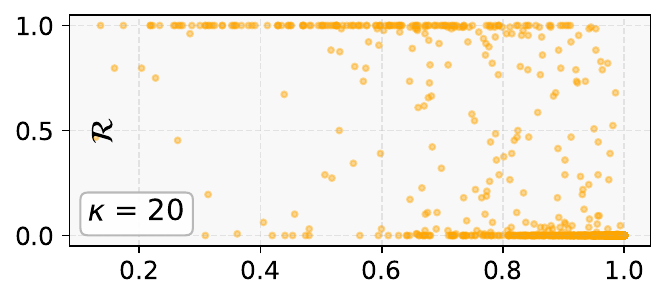}
    \end{subfigure}
    \begin{subfigure}{0.33\columnwidth}
        \centering
        \includegraphics[width=\textwidth]{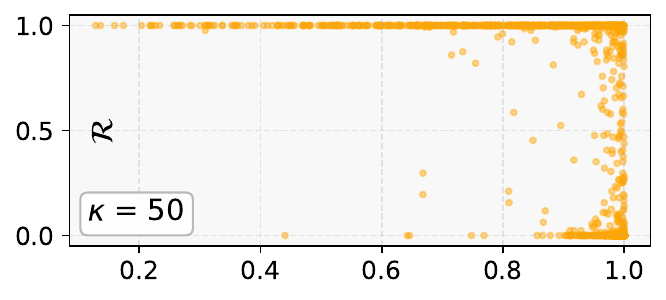}
    \end{subfigure}
    \caption{\small{WideResNet-101}}
    \end{subfigure}
    \begin{subfigure}{0.8\columnwidth}
    \begin{subfigure}{0.33\columnwidth}
        \centering
        \includegraphics[width=\textwidth]{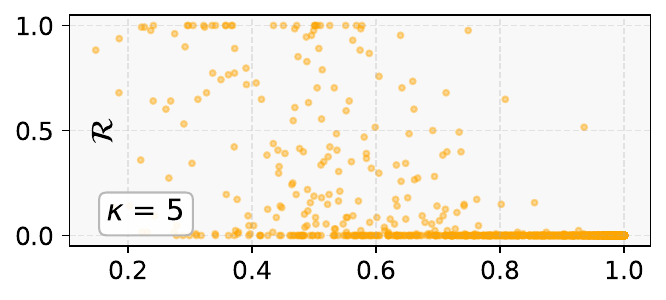}
    \end{subfigure}
    \begin{subfigure}{0.33\columnwidth}
        \centering
        \includegraphics[width=\textwidth]{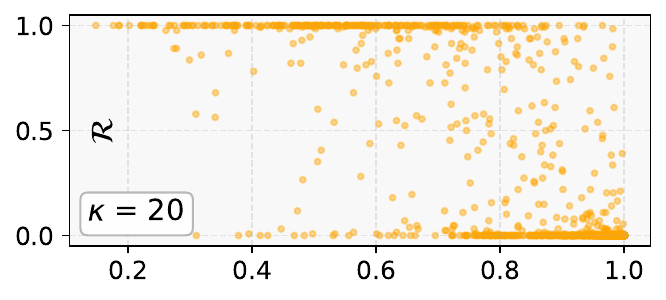}
    \end{subfigure}
    \begin{subfigure}{0.33\columnwidth}
        \centering
        \includegraphics[width=\textwidth]{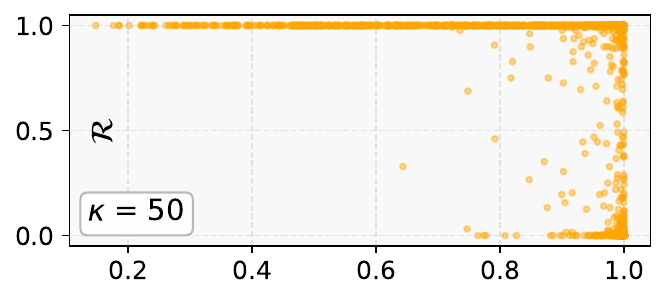}
    \end{subfigure}
    \caption{\small{VGG16}}
    \end{subfigure}
    \caption{\small{Additional results from the analysis in Figure \ref{fig:confidence_analysis}, with same settings when considering $\mathcal{R}$ for WideResNet101 and VGG-16. The x-axis reports the confidence, as the softmax probability of the predicted class.}}
    \label{fig:confidence_analysis_appendix}
\end{figure}

\section{Confidence Score Analysis}
\label{app:confidence_extra}

We further analyze the confidence-score trends introduced in
Section~\ref{sec:confidence_analysis} also for two additional models,  WideResNet-101 and VGG-16.
Figure~\ref{fig:confidence_analysis_appendix} reports the corresponding results.
Consistent with the results in the paper, we observe a clear relationship between directional perturbation risk and model confidence.
In particular, adversarial attacks that exhibit high perturbation risk values
(i.e., large $\mathcal{R}$) already at low $\kappa$ are strongly associated with lower prediction confidence.
This behavior remarks the trend and indicates that directions inducing statistically representative noisy failure regions tend to reduce model confidence more substantially than highly localized, worst-case perturbations, further supporting the proposed interpretation of the directional perturbation risk.

Interestingly, compared to the results reported in the main paper, although the overall trend is preserved across all tested models, ViT-B exhibits a more pronounced pattern.
Specifically, even for $\kappa > \kappa^*$ (e.g., $\kappa = 20$), no tested samples with high confidence (e.g., confidence $> 0.8$) achieve $\mathcal{R} < 0.5$. This observation highlights a particular reliability of transformer-based models to statistically representative perturbations and suggests that uncertainty estimates in such architectures may be especially informative from a robustness assessment perspective.

\end{document}